\documentclass[a4paper,fleqn]{cas-sc}

\usepackage[numbers]{natbib}
\usepackage{subcaption}
\usepackage{algorithm}
\usepackage{algorithmic}
\usepackage{tabularx}

\def\tsc#1{\csdef{#1}{\textsc{\lowercase{#1}}\xspace}}
\tsc{WGM}
\tsc{QE}

\begin{document}
\let\WriteBookmarks\relax
\def\floatpagepagefraction{1}
\def\textpagefraction{.001}

\shorttitle{}    

\shortauthors{}  

\title [mode = title]{Spatiotemporal Attention-Augmented Inverse Reinforcement Learning for Multi-Agent Task Allocation}  

\tnotemark[1]
\tnotetext[1]{This work was supported in part by the National Natural Science Foundation of China under Grant 62133011.}

\author[1]{Yin Huilin}[
    style=chinese,
    orcid=0000-0002-8507-015X
]
\cormark[1]
\ead{yinhuilin@tongji.edu.cn}
\credit{Conceptualization, Supervision, Funding acquisition, Writing -- review \& editing}
\author[1]{Yang Zhikun}[
    style=chinese,
    orcid=0009-0003-1555-0355
]
\ead{xiaoyuezk@tongji.edu.cn}
\credit{Conceptualization, Methodology, Software, Investigation, Validation, Writing -- original draft}

\author[2]{Zhang Linchuan}[
    style=chinese,
    orcid=0009-0008-3840-4648
]
\ead{jmzlc@tongji.edu.cn}
\credit{Writing -- review \& editing, Validation}

\author[3,4]{Daniel Watzenig}[
    orcid=0000-0002-5341-9708
]
\ead{daniel.watzenig@tugraz.at}
\credit{Supervision, Writing -- review \& editing}

\affiliation[1]{organization={College of Electronic and Information Engineering, Tongji University},
                addressline={No. 4800, Caoan Road},
                city={Shanghai},
                postcode={201804},
                state={Shanghai},
                country={China}}

\affiliation[2]{organization={Shanghai Research Institute for Intelligent Autonomous Systems, Tongji University},
                addressline={Building 17, Lane 55, Chuanhe Road},
                city={Shanghai},
                postcode={201210},
                state={Shanghai},
                country={China}}

\affiliation[3]{organization={Graz University of Technology},
                addressline={Rechbauerstraße 12},
                city={Graz},
                postcode={8010},
                state={Styria},
                country={Austria}}

\affiliation[4]{organization={Virtual Vehicle Research GmbH},
                addressline={Inffeldgasse 21A},
                city={Graz},
                postcode={8010},
                state={Styria},
                country={Austria}}

\cortext[cor1]{Corresponding author}

\begin{abstract}
Adversarial inverse reinforcement learning (IRL) offers a principled way to infer reward functions from expert demonstrations, yet its application in multi-agent task allocation (MATA) remains challenging due to non-stationary interactions, high-dimensional coordination, and unstable reward learning. In particular, unconstrained reward inference in adversarial multi-agent settings often suffers from high variance, poor credit assignment, and limited generalization. This paper proposes an attention-structured adversarial IRL framework for MATA that explicitly constrains reward inference through spatiotemporal representation learning. Multi-head self-attention (MHSA) is employed to model long-range temporal dependencies in inter-task agent trajectories, while graph attention networks (GAT) encode permutation-invariant relational structures between agents and tasks. Based on these representations, reward inference is formulated as a low-capacity, adaptive linear transformation of the environment reward, providing stable and interpretable guidance for multi-agent policy optimization. The proposed framework decouples reward inference from policy learning and optimizes the reward model adversarially against expert demonstrations, enabling effective preference extraction while mitigating reward instability in dynamic multi-agent environments. Extensive experiments on benchmark task allocation scenarios demonstrate that the proposed approach consistently outperforms representative multi-agent reinforcement learning (MARL) baselines across varying agent–task scales, achieving faster convergence, higher cumulative rewards, and improved spatial efficiency. The results indicate that attention-guided, capacity-constrained reward inference constitutes an effective neural computation mechanism for stabilizing adversarial IRL in multi-agent systems, offering a scalable and generalizable solution for complex task allocation problems.
\end{abstract}

\begin{graphicalabstract}
\includegraphics{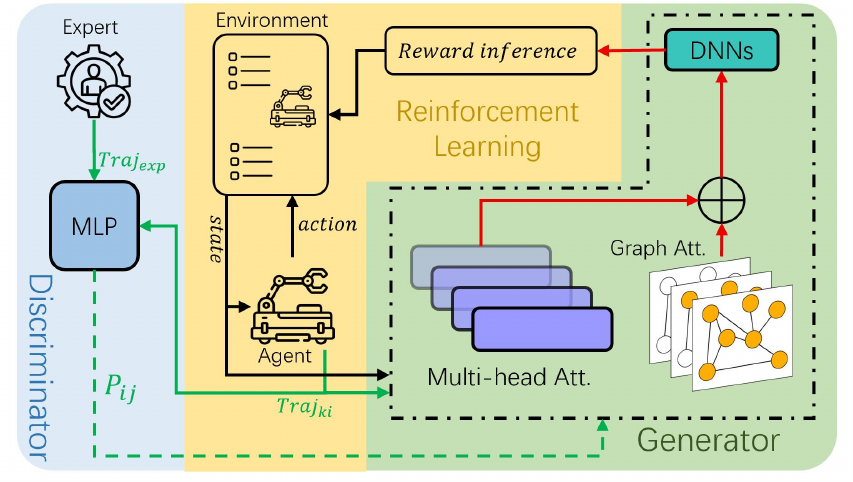}
\end{graphicalabstract}

\begin{highlights}
\item Proposes an attention-augmented inverse reinforcement learning framework for multi-agent task allocation.

\item Formulates reward inference as a capacity-constrained adaptive transformation guided by expert demonstrations.

\item Integrates temporal self-attention and graph attention to support structured reward inference in multi-agent systems.

\item Achieves improved reward stability, convergence efficiency, and coordination performance across diverse agent–task scales.

\end{highlights}

\begin{keywords}
Multi-agent systems \sep Multi-agent reinforcement learning \sep Inverse reinforcement learning \sep Spatiotemporal representation learning
\end{keywords}

\maketitle

\section{Introduction}
\label{intro}

With advances in automatic control, computing, and artificial intelligence, multi-agent systems (MAS) are increasingly deployed in industrial and public-service applications such as logistics, transportation, and search-and-rescue \citep{wang2022cooperative}. By coordinating multiple agents, MAS are able to address tasks whose workload, spatial scale, and temporal complexity exceed the capability of a single agent \citep{palmer2018modelling}, thereby providing the flexibility, fault tolerance, and robustness required by intelligent, application-driven systems \citep{ota2006multi}. A representative example is warehouse automation with automated guided vehicles (AGVs), where efficient material handling and distribution critically depend on coordinated multi-agent operation. In such systems, effective task allocation serves as a fundamental prerequisite for achieving high throughput and reliable service quality.

\begin{figure}
\centering \includegraphics[width=\textwidth]{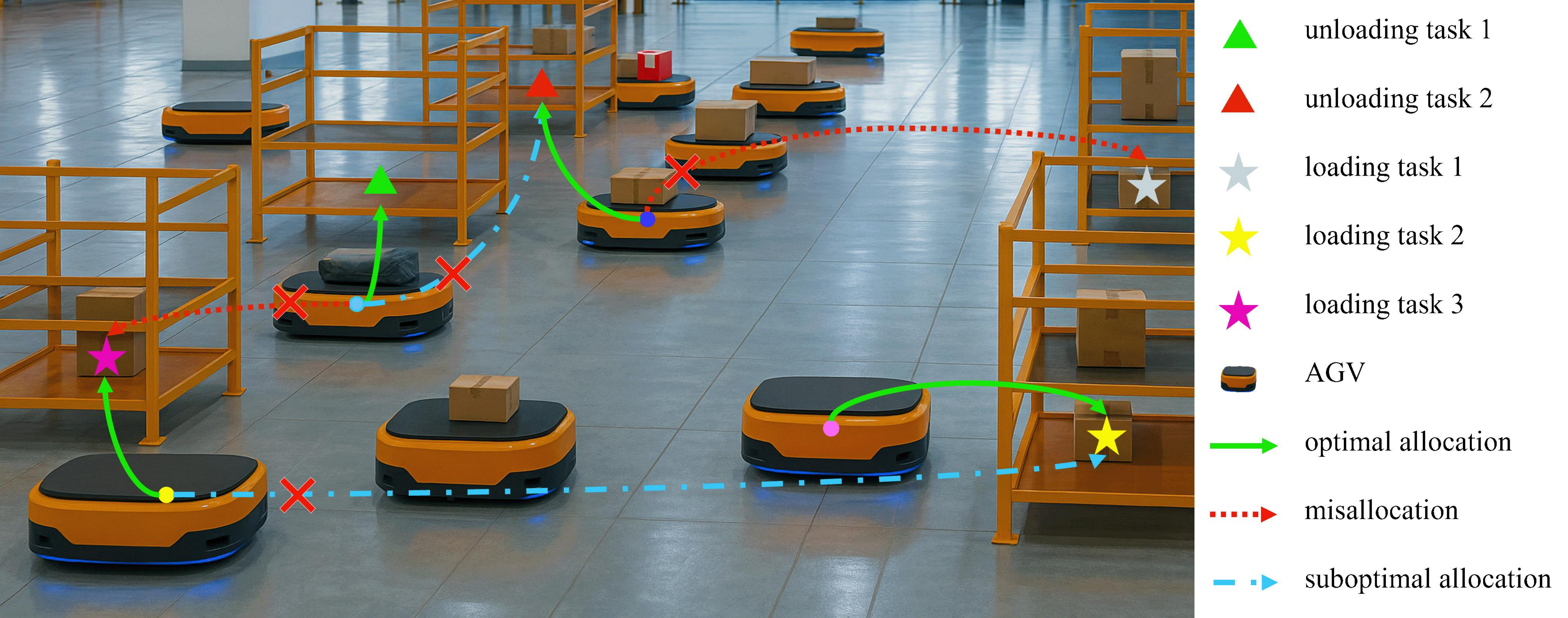} \caption{Schematic diagram of MATA for AGVs formation.}\label{agv} \end{figure}

As illustrated in Fig.~\ref{agv}, multi-agent task allocation (MATA) is a core decision-making problem whose solution quality directly determines system efficiency and resource utilization. When task complexity or workload outstrips the capacity of a single agent, cooperation among agents becomes indispensable \citep{skaltsis2023review}. Real-world deployments further introduce practical constraints, including tight timing requirements, uncertain task arrivals, and evolving spatial layouts, which demand task allocation methods that can scale and adapt online. Classical optimization-based approaches, such as G-CBBA \citep{kim2020bid} and Improved-QPSO \citep{zhang2020intelligent}, have demonstrated strong performance in small-scale and static environments. However, since MATA is NP-hard \citep{korsah2013comprehensive}, these methods often struggle to meet industrial requirements for scalability, adaptability, and real-time decision-making in large and dynamic settings \citep{skaltsis2021survey}. These limitations have motivated the growing adoption of learning-based approaches that can exploit data and interaction to support adaptive coordination in practice.

Among such approaches, deep reinforcement learning (DRL) has emerged as a promising paradigm for MATA by leveraging deep neural networks (DNNs) to approximate high-dimensional policies. Through repeated interaction with the environment, DRL enables agents to autonomously acquire decision-making strategies that adapt to uncertainty and operational variability \citep{SHAKYA2023120495}. Compared with traditional heuristics, DRL-based methods have shown improved performance in complex vehicular and logistics systems, yielding better coordination, throughput, and robustness \citep{seenu2020review}. Despite these advances, the practical deployment of DRL for large-scale MATA remains hindered by two persistent challenges. First, most DRL approaches rely on manually crafted reward functions, whose design is subjective, costly to adapt across scenarios, and often misaligned with real operational objectives. Second, exploration in high-dimensional multi-agent spaces is inherently inefficient, leading to slow convergence and unstable learning in the absence of strong priors \citep{dulac2021challenges}. These issues significantly limit the scalability and industrial applicability of DRL-based task allocation methods.

Inverse reinforcement learning (IRL) provides a complementary perspective by shifting the focus from reward specification to reward inference. By leveraging expert demonstrations, IRL aims to recover reward functions that encode implicit task preferences, thereby reducing reliance on handcrafted designs and aligning learning objectives with desired behaviors. From a knowledge-driven standpoint, expert demonstrations constitute structured prior information that can be incorporated into the learning process, enabling IRL-based methods to function as decision-support components for MATA. Moreover, such demonstrations can guide exploration and accelerate policy learning by constraining the search space. Classical IRL formulations, including Maximum Entropy IRL and Feature Expectation Matching, have demonstrated effectiveness in capturing implicit preferences in complex planning and control problems \citep{wu2020learning,shamsoshoara2024joint}.

Nevertheless, most existing IRL frameworks have primarily been developed for single-agent or relatively simple sequential decision-making tasks. Although they have achieved notable success in domains such as path planning and trajectory optimization \citep{wu2020learning,shamsoshoara2024joint}, their extension to complex multi-agent task allocation settings remains limited. In multi-agent environments, reward inference is complicated by non-stationary interactions, distributed decision-making, and intertwined agent behaviors, which collectively exacerbate the difficulty of learning stable and generalizable reward functions. To the best of our knowledge, IRL-based approaches have not yet been systematically investigated for large-scale, dynamic MATA problems, leaving an important methodological gap.

Furthermore, directly applying IRL to MATA is insufficient without addressing additional representational challenges. Conventional IRL formulations often lack the ability to capture long-range temporal dependencies in agent trajectories, as well as the global coordination structures arising from interactions among agents and tasks. In multi-agent task allocation, effective reward inference requires representations that can encode both temporal context and relational structure, while providing inductive bias for credit assignment across time and agents. Attention mechanisms offer a natural means of addressing these requirements. In particular, multi-head self-attention (MHSA) is well suited for modeling temporal dependencies and contextual correlations in sequential behaviors, whereas graph attention networks (GAT) enable structured representation of agent–task interactions in a permutation-invariant manner.

Motivated by these observations, this paper proposes a spatiotemporal fusion IRL framework for MATA that integrates MHSA and GAT to support reward inference in multi-agent systems. MHSA is employed to capture long-range temporal correlations in inter-task agent trajectories, while GAT leverages graph-structured representations to model global coordination patterns between agents and tasks. By combining these mechanisms, the proposed framework learns more expressive and structured representations for reward inference, resulting in adaptive reward guidance that improves coordination efficiency, scalability, and robustness in dynamic MATA scenarios.

The primary contributions of this study are summarized as follows:
\begin{enumerate}
\item We propose an IRL-based framework for multi-agent task allocation that infers adaptive reward signals from expert demonstrations, reducing dependence on manually designed reward functions and facilitating flexible coordination in dynamic environments.
\item We introduce an attention-augmented representation for reward inference in multi-agent systems, in which MHSA captures temporal dependencies in agent trajectories and GAT models agent–task interactions and global coordination structures.
\item We conduct extensive experiments across diverse agent–task scales to evaluate the proposed framework, demonstrating consistent improvements over representative multi-agent reinforcement learning baselines in terms of cumulative reward, task completion efficiency, and travel distance.
\end{enumerate}

\section{Related Work}
\label{sec:related_work}

This section reviews existing studies on multi-agent task allocation (MATA), with an emphasis on how coordination strategies and learning mechanisms have evolved from optimization-based formulations to data-driven approaches. The discussion is organized into four categories: (1) traditional MATA methods and their limitations, (2) deep reinforcement learning (DRL)-based approaches for scalable coordination, (3) inverse reinforcement learning (IRL) methods for reward inference, and (4) attention mechanisms for modeling complex dependencies. This progression highlights the limitations of existing paradigms and motivates the integration of structured reward inference with spatiotemporal representation learning.

\subsection{Traditional MATA Methods}

Early approaches to MATA primarily rely on optimization-based methods and heuristic algorithms. Auction-based strategies such as G-CBBA extend consensus-based bundle algorithms to enable multi-agent coordination through iterative bidding mechanisms. \cite{LIU2025127299} further propose a hybrid genetic algorithm tailored to multi-robot task allocation with limited span, demonstrating improved efficiency in minimizing cycle time on industrial benchmarks. While these methods are effective in small-scale and static environments with known task distributions, they exhibit fundamental limitations in real-world deployments. As MATA is NP-hard, the computational complexity of optimization-based approaches grows rapidly with problem size, making them unsuitable for large-scale scenarios. Moreover, such methods typically require complete re-computation when task arrivals or execution conditions change, limiting their adaptability in dynamic environments \citep{skaltsis2021survey}. These challenges have driven the shift toward learning-based approaches that can exploit experience to support adaptive coordination.

\subsection{DRL-based MATA Methods}

Deep reinforcement learning has been increasingly adopted for MATA due to its ability to approximate high-dimensional policies and adapt through interaction. Based on training and execution paradigms, existing DRL-based MATA methods can be categorized into three frameworks \citep{gronauer2022multi}.

\textbf{Centralized Training and Centralized Execution (CTCE):} Recent studies demonstrate the effectiveness of centralized DRL for large-scale combinatorial routing problems. For example, Yin and Yang propose a hierarchical deep reinforcement learning framework that decomposes surface mount vehicle routing into assignment and sequencing subproblems, where attention-based encoder–decoder networks autonomously learn assignment rules without manually designed heuristics \citep{yin2026hdr}. While such CTCE frameworks achieve strong performance in structured industrial settings, they rely on centralized optimization and remain sensitive to problem scale and environmental changes.

\textbf{Centralized Training and Decentralized Execution (CTDE):} CTDE frameworks enable distributed execution after centralized training and thus offer improved scalability. Representative methods include CapAM \citep{paul2022learning}, which combines capsule networks with attention-based GNNs, auction-based RL frameworks \citep{liu2022multi}, and hierarchical decomposition approaches based on QMIX \citep{wang2023hcta}. More recent studies have explored swarm coordination \citep{lv2024local}, conflict-free vehicle assignment \citep{ratnabala2025magnnet}, and UAV-assisted crowd sensing \citep{liu2024coverage}. While CTDE methods enhance robustness, they often require extensive communication during training and remain sensitive to non-stationary agent interactions.

\textbf{Decentralized Training and Decentralized Execution (DTDE):} Fully decentralized approaches such as HIPPO-MAT \citep{ratnabala2025hippo} and GAPPO \citep{fang2025efficient} eliminate centralized coordination by combining local policy learning with evolutionary or graph-based mechanisms. Although DTDE methods perform well under communication constraints, they frequently converge to suboptimal solutions due to limited global information and weak coordination priors.

Despite these advances, recent studies indicate that improving exploration efficiency alone is insufficient to fully address coordination challenges in multi-agent systems. For instance, Cao et al. propose LECMARL, which integrates lazy mechanisms and efficient exploration strategies to mitigate sparse-reward issues in cooperative MARL \citep{CAO2026132578}. Although such intrinsic-motivation-based designs significantly accelerate learning, they still operate under manually specified reward structures and do not infer task-level coordination objectives from expert behavior.

\subsection{Studies of IRL Methods}

Inverse reinforcement learning addresses the reward design challenge by inferring reward functions directly from expert demonstrations, thereby reducing manual specification while aligning learning objectives with observed behaviors. In single-agent domains, IRL has achieved notable success in path planning \citep{lian2022data}, joint optimization \citep{shamsoshoara2024joint}, and related sequential decision-making problems. Maximum entropy IRL variants introduce principled regularization by incorporating structural priors, such as submodular reward functions \citep{wu2020learning} and coverage-based exploration strategies \citep{wu2022adaptive}. Recent advances further integrate IRL with deep architectures, including hierarchical transformers for motion forecasting \citep{azadani2024hierarchical}, behavior cloning under non-stationary dynamics \citep{sivakumar2024inverse}, and query-based offline learning frameworks \citep{sun2023query}. Adversarial IRL formulations, such as Generative Adversarial Imitation Learning (GAIL) \citep{ho2016generative} and Adversarial Inverse Reinforcement Learning (AIRL) \citep{fu2018learningrobustrewardsadversarial}, bridge IRL and generative modeling by matching expert and learner occupancy measures while enabling scalable reward learning.

Extensions of IRL to multi-agent settings remain comparatively limited. MIRL \citep{freihaut2025feasiblerewardsmultiagentinverse} introduces entropy-regularized game-theoretic formulations to ensure equilibrium uniqueness, while \cite{waelchli2023discovering} recover individual reward functions from collective behaviors using off-policy learning. Game-theoretic approaches \citep{goktas2025efficient} further frame MIRL as a generative-adversarial optimization problem with polynomial-time solutions. However, most existing multi-agent IRL methods focus on homogeneous agents and fully observable environments, and they do not explicitly address the representational and stability challenges arising in large-scale, dynamic MATA scenarios. In particular, how reward inference can be structured to remain stable under non-stationary multi-agent interactions remains an open problem.

\subsection{Attention Mechanisms in MAS}

Attention mechanisms have emerged as powerful tools for capturing complex dependencies in multi-agent systems. Multi-head self-attention (MHSA) has been widely used to model temporal correlations in sequential decision-making, enabling agents to selectively attend to relevant historical information \citep{seenu2020review}. Graph attention networks (GAT) provide a natural representation for agent–task relationships by encoding interactions as weighted graphs, thereby capturing both local neighborhoods and global coordination patterns. Attention mechanisms have already been incorporated into MATA methods, with CapAM being a representative example. Nevertheless, existing studies predominantly employ attention for policy learning or action selection. Its potential role as a structural inductive bias for reward inference, particularly in conjunction with IRL, remains largely unexplored.

\subsection{Summary and Research Gap}

In summary, existing approaches to multi-agent task allocation span optimization-based methods, DRL-based coordination strategies, and IRL-based reward inference frameworks. While each paradigm addresses specific challenges, none fully satisfies the requirements of industrial-scale MATA in terms of scalability, adaptability, and alignment with operational objectives. In particular, three key challenges remain insufficiently addressed: (1) reward signals that capture complex multi-agent coordination objectives are rarely inferred automatically, (2) long-range temporal dependencies in agent behaviors are insufficiently modeled in dynamic task allocation settings, and (3) global agent–task interaction structures are not effectively integrated into reward learning frameworks.

These gaps motivate the development of methods that jointly address reward inference and representation learning in multi-agent systems. To this end, this work proposes an attention-augmented IRL framework that integrates temporal and relational attention mechanisms to support structured and stable reward inference for MATA.

\section{System Model and Problem Formulation}
\label{sec:system_model}
To support efficient and conflict-free task allocation in the MATA setting, key assumptions and constraints on agent behavior, task timing, and energy consumption are defined in the system model. Based on these formulations, an optimization problem is constructed to capture the trade-offs between energy usage and task completion efficiency.
\subsection{System Model}
This study considers a MATA scenario common in logistics and emergency response, featuring multi-task agents with single-task assignment and instantaneous allocation. In this scenario, each task requires only one agent for completion and is removed upon completion. Agents perform multiple tasks sequentially until all tasks are completed. No priority or mandatory dependency exists among tasks. 

In the MATA framework, the agent set of \(N\) agents is denoted as \(\Lambda = \{ \lambda_1, \lambda_2, \ldots,\lambda_i,\ldots, \lambda_N \}\), and the task set of \(M\) tasks is denoted as \(J = \{ j_1, j_2, \ldots, j_k,\ldots, j_M \}\), where \(N < M, i<N, k<M\). All tasks are predefined and remain fixed during execution. The agent set is homogeneous, meaning each agent can complete any task. This design avoids the complexity of handling heterogeneous capabilities and ensures an equal opportunity for task assignment among agents. Task allocation is represented by \( C : J \to \Lambda \). If task \( j_k \) is assigned to agent \( \lambda_i \), then \( C_{ki} = 1 \); otherwise, \( C_{ki} = 0 \).

The following notations are used: \( Z = \{z_1, z_2, \dots, z_M\} \) for task arrival times, \( O = \{o_1, o_2, \dots, o_M\} \) for task waiting times, and \( G = \{g_1, g_2, \dots, g_M\} \) for task durations. Although our scenario adopts instantaneous allocation, these temporal variables are defined in a general form to maintain compatibility with settings involving release delays or nonzero waiting. For any two tasks \( j_u \) and \( j_v \) (\( u \neq v \)) assigned to the same agent, the combined end time of one task must not interfere with the start time of the other. This requires that either \( z_u + o_u + g_u \leq z_v + o_v\), meaning task \( j_u \) finishes before task \( j_v \) starts, or \( z_v + o_v + g_v \leq z_u + o_u\), meaning task \( j_v \) finishes before task \( j_u \) starts. Either condition ensures sequential task handling without time conflicts.

With a length of \( L_{ki} \) as the trajectory traversed by agent \( \lambda_i \), the sequence of coordinate pairs \( Traj_{ki} \) is defined as \( [(x_1, y_1), (x_2, y_2), (x_3, y_3), \dots, (x_{L_{ki}}, y_{L_{ki}})] \),  while transitioning from the completion of its previous task to the execution of task \( j_k \). Then the moving distance is \( \sum_{d=1}^{L_{ki}-1} \left\| Traj_{ki}^{(d+1)} - Traj_{ki}^{(d)} \right\|_2 \).

The following performance metrics are used to evaluate MATA performance:
\begin{equation}
\label{min_travel_distance}
\min \sum_{j_k \in J} \sum_{\lambda_i \in \Lambda} 
C_{ki} \sum_{d=1}^{L_{ki}-1} 
\left\|Traj_{ki}^{(d+1)} - Traj_{ki}^{(d)} \right\|_2 .
\end{equation}

\begin{equation}
\label{min_waiting_time}
\min \sum_{j_k \in J} o_k .
\end{equation}

Metrics (\ref{min_travel_distance}) and (\ref{min_waiting_time}) represent minimizing travel distance and minimizing waiting time respectively. These indicators reflect different aspects of efficiency. Their importance varies depending on the application.

\subsection{Problem Formulation}
Let \( t_{\text{total}} = \max_{j_{i} \in J}(z_i + o_i + g_i) - \min_{j_{k} \in J}(z_k) \) represent the total elapsed time since task allocation began. Let the total energy of each robot be \( E \), which decreases over time. The energy consumed during movement is denoted as \( E_c \), and the energy consumed to complete task \( j_k\in J \) is denoted as \( e_k \). To simplify the constraint formula, we introduce the following auxiliary variables: \( \theta_{uv} = \sum_{\lambda_i \in \Lambda} C_{ui} C_{vi} \), \( \rho_k = z_k + o_k + g_k \). \(\theta_{uv}\) is an indicator variable that denotes whether tasks \(j_u\) and \(j_v\) are assigned to the same agent. Here, $\rho_k$ represents the completion time of the task \( j_k\).

The problem is defined as the following mixed-constraint optimization problem:
\vspace{-3pt}
\begin{subequations}
\label{eq_3}
\begin{gather}
\min \quad \alpha \sum_{\lambda_i \in \Lambda} \left( E_c + \sum_{j_k \in J} e_k \cdot C_{ki} \right) + \beta t_{\text{total}} \label{eq_4a} \\
\text{s.t.} \sum_{\lambda_i \in \Lambda} C_{ki} = 1, \quad \forall j_k \in J \label{eq_4b} \\
\begin{aligned}
    & \quad \forall u, v \in \{1, 2, \dots, M\}, \, u \neq v : \\
    & \theta_{uv} = 1 \; \Rightarrow \; \bigl(\rho_u \le z_v + o_v \;\lor\; \rho_v \le z_u + o_u\bigr)
\end{aligned} \label{eq_4c} \\
E - E_c - \sum_{j_k \in J} e_k \cdot C_{ki} \geq 0, \quad \forall \lambda_i \in \Lambda \label{eq_4d} ,
\end{gather}
\end{subequations}

\noindent where \( \alpha \) and \( \beta \) are non-negative coefficients. Objective function (\ref{eq_4a}) aims to balance two key objectives: maximizing remaining energy and minimizing task completion time. Maximizing the remaining energy after task completion extends system operation time and enhances resource utilization, while minimizing total completion time ensures task execution efficiency. The weights of these objectives are controlled by \( \alpha \) and \( \beta \), where \( \alpha \) emphasizes energy conservation and \( \beta \) emphasizes time efficiency. These weights can be adjusted based on application scenarios to optimize performance.

Constraint (\ref{eq_4b}) ensures that each task \( j_k \in J \) must be assigned to exactly one agent \( \lambda_i \in \Lambda \), preventing tasks from being omitted or redundantly assigned. This constraint reflects the physical requirement that each task is a single-agent task requiring a clear and unique responsibility allocation. Constraint (\ref{eq_4c}) ensures that each agent can only perform one task at any given time, avoiding scheduling conflicts. If two tasks \( j_u \) and \( j_v \) are assigned to the same agent, their execution times must not overlap. Constraint (\ref{eq_4d}) limits the energy consumption of agents. The remaining energy of each agent is the initial total energy \( E \) minus the total energy consumed in performing tasks. The remaining energy is required to be non-negative to ensure that the agent will not fail due to energy exhaustion during task execution.

\section{Proposed Approaches}

In this section, the system model and problem formulation introduced in Section \ref{sec:system_model} are reformulated as a Markov Decision Process (MDP). Building on this framework, an efficient algorithm based on IRL is developed to effectively address the MATA.
\subsection{DRL based Task Allocation}
In general, reinforcement learning is modeled as a sequential decision-making process based on the framework of a Markov decision process (MDP). An MDP is typically represented as a tuple $\langle S, A, T, R, \gamma \rangle$, where the state at time $t+1$ depends only on the state at time $t$ and is independent of all previous states, thereby satisfying the Markov property. In this formulation, $S$ denotes the finite state space and $A$ denotes the finite action space. The transition dynamics are described by the function $T: A \times S \to S$, which specifies the probability of reaching the next state $s_{t+1}$ from the current state $s_t$ when action $a_t$ is taken. The reward function $R: S \times A \to \mathbb{R}$ assigns the expected reward $r_{t+1}$ obtained after executing action $a_t$ in state $s_t$. The discount factor $\gamma \in [0,1)$ determines the present value of future rewards by applying an exponential decay over time. A policy $\pi$ is defined as a computable mapping from states to actions, $\pi: S \to A$, and represents the agent's decision-making strategy within the MDP framework.

In this paper, it is assumed that each agent independently determines its own actions to maximize its discounted future rewards. The immediate reward obtained by each agent during the training process depends on its position and coordination. The designs of the state space \( S \), action space \( A \) and reward function \( R \) are as follows:

State Space \( S \): In smart transportation and vehicular logistics scenarios, an individual agent's perception via vision or LiDAR is limited. However, reliable communication enables agents to share global state information, including the positions of agents and tasks, ensuring equal access to a comprehensive state representation. At timestep $t$, the system state is represented by the positions and completion information of all agents and tasks:
$s_t = \left\{ \mathbf{p}_{1,t}, \ldots, \mathbf{p}_{N,t}, \mathbf{q}_{1,t}, \ldots, \mathbf{q}_{M,t} \right\}$,
where $\mathbf{p}_{i,t} \in \mathbb{R}^3$ denotes the state vector of agent $\lambda_i$, and $\mathbf{q}_{j,t} \in \mathbb{R}^3$ denotes the state vector of task $j$. 
The first two dimensions of each vector represent the two-dimensional position, while the third dimension indicates the completion status of the corresponding agent or task. 

Action Space \( A \): As homogeneous agents with identical characteristics, all agents share the same action space, consisting of movement distance and direction. To simplify the problem, a uniform movement speed can be assumed, reducing the action space to movement direction only. The joint action of all agents is denoted as \( \mathbf{a}_t = (a_{1,t}, a_{2,t}, \ldots, a_{N,t}) \), with corresponding joint policies \( \boldsymbol{\pi}_t = (\pi_{1,t}, \pi_{2,t}, \ldots, \pi_{N,t}) \).  

Reward Function \( R \): Based on the objective function (\ref{eq_4a}), the reward balances energy consumption and task efficiency. Assuming uniform energy consumption per unit distance, let \( d_{i,t} \) denote the distance moved by agent \( \lambda_i \) at time step \( t \). The corresponding energy consumption is given by \( E_c = f(d_{i,t}) \), where \( f(*) \) maps distance to energy cost. Although \( E_c \) reflects actual energy use, it is often difficult to measure precisely in practice, whereas distance \( d_{i,t} \) and time \( t \) are readily observable. Therefore, the reward function is then formulated as a weighted combination of energy consumption with coefficient \(\eta\) and task efficiency with coefficient \(\phi\):
\begin{equation}
\label{reward}
r_{i,t} = \eta f(d_{i,t}) + \phi t .
\end{equation}
In multi-agent DRL, the expected return \( J_i(\pi_i) \) of each agent is maximized  based on the maximum entropy principle $
J_i(\pi_i) = \mathbb{E}_{\tau \sim \boldsymbol{\pi}} \left[ \sum_{t=0}^\infty \gamma^t r_{i,t}(s_t, \mathbf{a}_t)  \ \right]$, where \( r_{i,t}(s_t, \mathbf{a}_t) \) represents the immediate reward for agent \( \lambda_i \). Since the state transition is determined by the joint action \( \mathbf{a}_t \), the reward for a single agent also depends on the joint action. Multi-agent DRL includes the action-value network and policy network. The action-value function \( Q_i(s_{t}, \mathbf{a}_{t}) \) for each agent \( \lambda_i \) is defined as $Q_i(s_t, \mathbf{a}_t) = \mathbb{E}_{\tau \sim \boldsymbol{\pi}} \left[ \sum_{k=0}^{\infty} \gamma^k r_{i,t}(s_{t+k}, a_{i,t+k}) \middle| s_t, \mathbf{a}_t \right] $. The Bellman error is used as the objective function of action-value networks:

\begin{equation}
\label{loss_Q}
\mathcal{L}_Q = \mathbb{E}_{s_t, \mathbf{a}_t, r_t, s_{t+1}} \left[ \left( Q_i(s_t, \mathbf{a}_t) - y_t \right)^2 \right] ,
\end{equation}

\noindent where  \( y_t = r_t + \gamma \mathbb{E}_{\mathbf{a}_{t+1}} \left[ Q_i(s_{t+1}, \mathbf{a}_{t+1}) \right]
 \) is the target value. Policy optimization is usually based on maximizing the policy gradient of the $Q$ value:

\begin{equation}
\label{loss_pi}
\nabla_{\theta_i} J(\pi_i) = \mathbb{E}_{s_t , a_{i,t} \sim \pi_i} \left[ \nabla_{\theta_i} \log \pi_i(a_{i,t} | s_t) Q_i(s_t, \mathbf{a}_t) \right] .
\end{equation}

In the remainder, the MARL optimizer is used as a fixed backend. We decouple reward inference from expert demonstrations and spatiotemporal representation learning. The former provides adaptive reward guidance during training, while the latter constructs the representations consumed by the reward module and the MARL policy.

\subsection{Adversarial Reward Inference Framework}
\label{FRAME}
The manual design of reward functions in MATA is often constrained by conflicting objectives and subjective assumptions, which can limit adaptability and lead to suboptimal policies in high-dimensional environments \citep{arora2021survey}. To address these challenges, we propose an IRL-based framework that leverages expert demonstrations as structured priors for data-driven reward adaptation. Drawing on the adversarial learning principles of GAIL and AIRL, our approach adopts adversarial reward inference as a training paradigm. Instead of aiming for direct policy imitation or exact reward recovery, the proposed framework employs a generator-discriminator architecture to infer adaptive reward signals that align multi-agent policy optimization with implicit expert preferences, thereby reducing subjectivity and enhancing robustness in dynamic scenarios.

\begin{figure}
\centering
\includegraphics[width=.9\columnwidth]{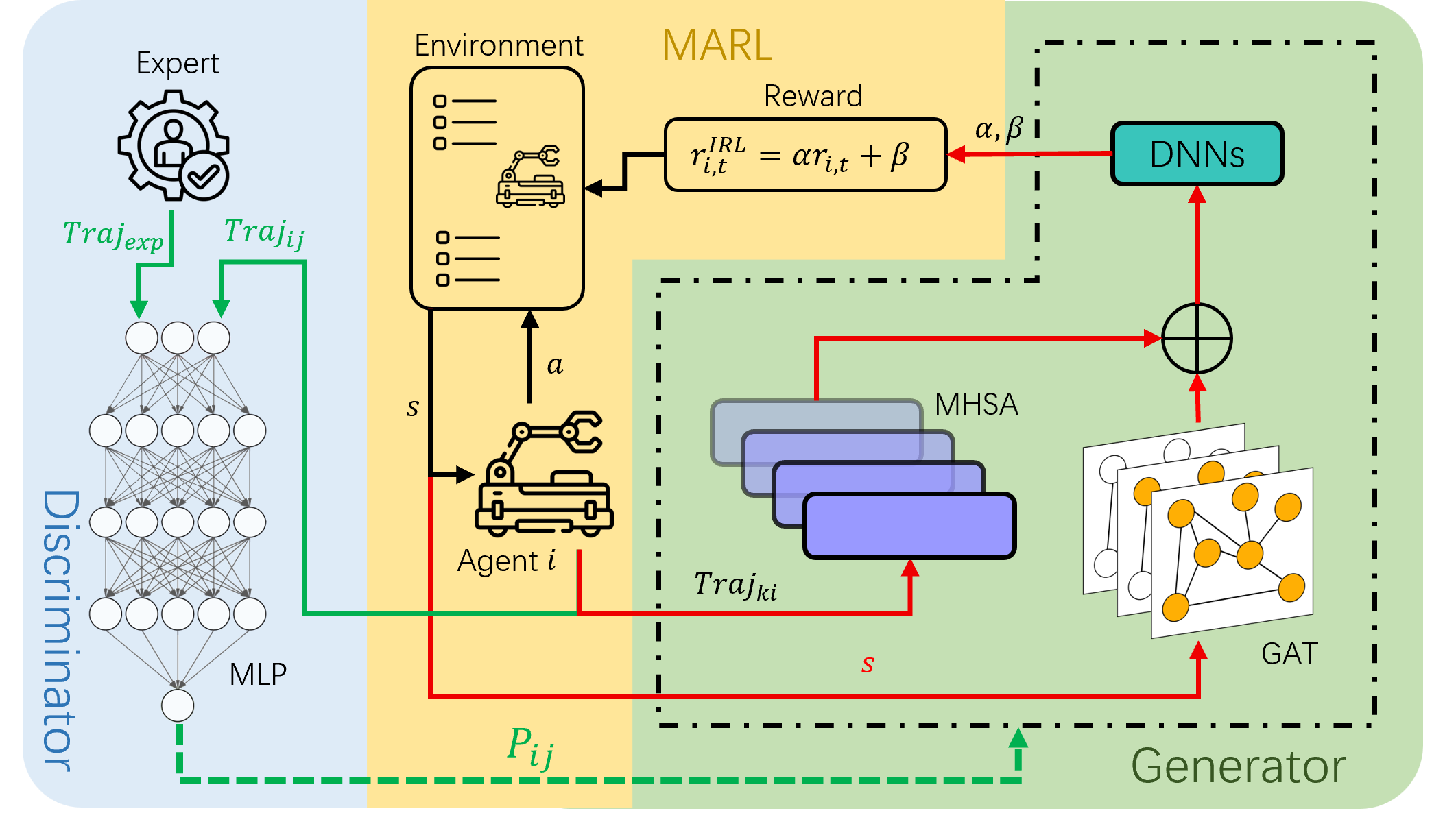}
\caption{Overall framework of the proposed adversarial reward inference approach for MATA. The MARL backend (yellow) interacts with the environment using adapted rewards produced by the generator (green), while the discriminator (blue) provides adversarial feedback based on expert demonstrations.}
\label{framwork}
\end{figure}

Unlike approaches that directly apply adversarial imitation learning to optimize policy networks, our framework applies adversarial learning at the reward inference level, which indirectly guides policy optimization through a standard multi-agent reinforcement learning backend. This design alleviates limitations of direct imitation, including restricted generalization in expanded state spaces and limited interpretability of learned behaviors. By focusing on reward guidance instead of policy cloning, the inferred reward signals provide an intuitive measure of action quality differences and facilitate analysis and optimization. Moreover, since policy learning is not constrained to strict imitation of expert trajectories, the framework may enable continued performance improvement beyond the demonstrated behaviors.

The overall framework is illustrated in Fig.~\ref{framwork}. The proposed framework consists of a MARL backend and an IRL-based reward inference module. The yellow-shaded area represents the MARL process, including the environment, agents, and policy optimization. The green-shaded area denotes the reward inference pipeline, which comprises a representation encoder and a reward adaptation module. Specifically, MHSA and GAT are employed to construct spatiotemporal representations of agent trajectories and agent–task interactions. These representations are then fed into a lightweight neural network to produce adaptive reward coefficients $\alpha$ and $\beta$, which are used to revise the original reward during training. Importantly, the representation encoder and the reward inference module are conceptually decoupled, and alternative encoders can be incorporated without altering the overall framework. The blue-shaded area represents the discriminator, which distinguishes expert demonstrations from policy-generated trajectories and provides adversarial feedback for optimizing the reward inference module. Black arrows indicate state transitions and action executions in the MARL loop. Green arrows represent expert demonstrations and discriminator outputs that serve as supervisory signals for reward inference. Red arrows illustrate information flow within the representation encoder, highlighting the integration of temporal and structural features.

In the proposed framework, reward inference is performed at the inter-task trajectory level with additional state context, where expert demonstrations provide supervisory signals for guiding reward adaptation during training. For each agent $\lambda_i$, the trajectory segment between two consecutive tasks, together with the corresponding state information, constitutes the basic input unit for reward inference. Rather than operating directly on raw state-action pairs, the framework maps this joint input into a compact representation space.

Specifically, a joint representation \(f_{ki} = \phi(Traj_{ki}, s)\) is constructed, where $\phi(\cdot)$ denotes a representation encoder that summarizes motion patterns from inter-task trajectories and contextual information from states. The design of $\phi(\cdot)$ is independent of the reward inference objective and will be specified in Section \ref{GIFE}.

Based on the joint representation $f_{ki}$, the reward adaptation module employs a lightweight learnable mapping, instantiated as the DNNs block in Fig.~\ref{framwork}, to produce two adaptive coefficients, $\alpha_{ki}$ and $\beta_{ki}$. As the module operates on compact representations and outputs only low-dimensional scaling and shifting terms, a limited-capacity formulation is sufficient and helps maintain training stability. The adapted reward for agent $\lambda_i$ at timestep $t$ is defined as

\begin{equation}
r^{\text{IRL}}_{i,t} = \alpha_{ki} \, r_{i,t} + \beta_{ki},
\end{equation}
where $r_{i,t}$ denotes the reward provided by the environment.

Unlike many existing methods \cite{sivakumar2024inverse}, this study retains the use of manually designed functions. This approach provides the neural network with an initial reasonable direction, reducing the time spent exploring unnecessary strategies during the learning process. Additionally, it offers a stable baseline, reducing dependency on data quality and enhancing system robustness. The linear form is adopted for three complementary reasons. First, it preserves interpretability, as $ \alpha_{ki}$ and $\beta_{ki}$ serve as explicit scaling and shifting terms. Second, it reduces the risk of overfitting compared with highly expressive nonlinear mappings, thereby maintaining the semantic structure of the handcrafted reward. Third, it produces stable, low-variance updates during training, which facilitates reliable convergence while still allowing flexible adaptation of the reward signal.

In practice, although the reward adaptation coefficients are inferred based on inter-task trajectory segments and state context, we adopt a simplified formulation by using shared coefficients $\alpha$ and $\beta$ during training: 

\begin{equation}
r^{\text{IRL}}_{i,t} = \alpha\, r_{i,t} + \beta,
\end{equation}

First, reflecting the homogeneity of the agents, this formulation enforces permutation invariance, ensuring consistent reward guidance regardless of agent identity. This prevents excessive variability at the step level which may otherwise lead to unstable policy optimization in multi-agent settings. Second, since reward adaptation serves as a form of global guidance rather than fine-grained reward reconstruction, a compact parameterization is sufficient to capture relative preference trends inferred from expert demonstrations. This simplification also improves computational efficiency and enhances the interpretability of the adapted reward.

For convenience of exposition under the adversarial learning framework, we collectively refer to the representation encoder and the reward adaptation module as the generator. It is important to note that, in this context, the generator does not synthesize actions or trajectories, nor does it directly optimize the policy network. Instead, it produces adaptive reward guidance based on trajectory and state representations, and is optimized adversarially against the discriminator.

In the adversarial optimization phase, as illustrated in the blue-shaded area of Fig.~\ref{framwork}, an $n$-layer multilayer perceptron (MLP) is employed as the discriminator. The discriminator is trained using binary cross-entropy loss with source labels indicating the origin of each trajectory sample: expert-provided trajectory segments are assigned label 1, while policy-generated trajectory segments are assigned label 0. To accommodate the fixed input requirement of the discriminator, inter-task trajectory segments are normalized via uniform resampling and concatenated with auxiliary features to form the fixed-length representation $\tau$ prior to MLP evaluation. This preprocessing serves purely as an interface operation and is independent of the reward inference mechanism. The discriminator outputs a scalar score that reflects the consistency of the input with expert demonstrations:

\begin{equation}
\label{eq_15}
P_{ki} = \sigma \left( W_n \sigma \left( W_{n-1} \dots \left( W_1 \tau + b_1 \right) + b_{n-1} \right) + b_n \right) ,
\end{equation}

\noindent where $\sigma(\cdot)$ denotes a non-linear activation function, and $W_i$ and $b_i$ represent the weight matrix and bias vector of the $i$-th layer, respectively.

Under the adversarial framework, the generator—comprising the representation encoder and the reward adaptation module—is optimized to produce reward guidance that increases the discriminator-assigned consistency scores of policy-generated trajectory segments. The loss function for the generator is defined as

\begin{equation}
\label{loss_gen}
\mathcal{L}_{\text{gen}} = -\mathbb{E}_{\mathrm{Traj}_{ki}} \left[ \log P_{ki} \right] .
\end{equation}

The discriminator is optimized to distinguish expert-generated and policy-generated trajectory segments by minimizing the following loss:

\begin{equation}
\label{loss_dis}
\mathcal{L}_{\text{disc}} =
- \mathbb{E}_{\mathrm{Traj}_{\text{exp}}} \left[ \log P_{\text{exp}} \right]
- \mathbb{E}_{\mathrm{Traj}_{ki}} \left[ \log (1 - P_{ki}) \right] .
\end{equation}

Through minimizing the binary cross-entropy loss, the discriminator learns to distinguish expert-consistent trajectory segments from policy-generated ones in the representation space. The resulting adversarial feedback serves as a learning signal for the generator, guiding it to adjust reward guidance such that policy-generated trajectories become increasingly consistent with expert demonstrations. Importantly, this adversarial interaction is used solely to optimize reward inference and does not directly update the policy network.
\subsection{Attention-Augmented Representation}
\label{GIFE}

Effective reward inference in MATA critically depends on the quality of spatiotemporal representations extracted from inter-task trajectories and state context. In contrast to single-agent or static task settings, MATA involves dynamic agent mobility, evolving task assignments, and non-stationary interactions among agents and tasks, which jointly induce long-range temporal dependencies and complex relational structures. To capture these characteristics, the representation encoder $\phi(\cdot)$ must simultaneously model temporal correlations along inter-task trajectories and structural dependencies within the agent–task system.

To address these challenges, we design a spatiotemporal representation encoder that instantiates $\phi(\cdot)$ using MHSA and GAT. MHSA is employed to capture temporal dependencies and contextual correlations within inter-task trajectory segments, while GAT operates on graph-structured representations to model interactions among agents and tasks. This design allows the encoder to extract compact, expressive features that are well aligned with the subsequent reward inference objective, while remaining decoupled from the adversarial learning framework described in Section \ref{FRAME}.

\begin{figure}
    \centering
    \hspace{0.3cm}
    \begin{subfigure}[b]{0.28\textwidth}
        \centering
        \includegraphics[width=\textwidth]{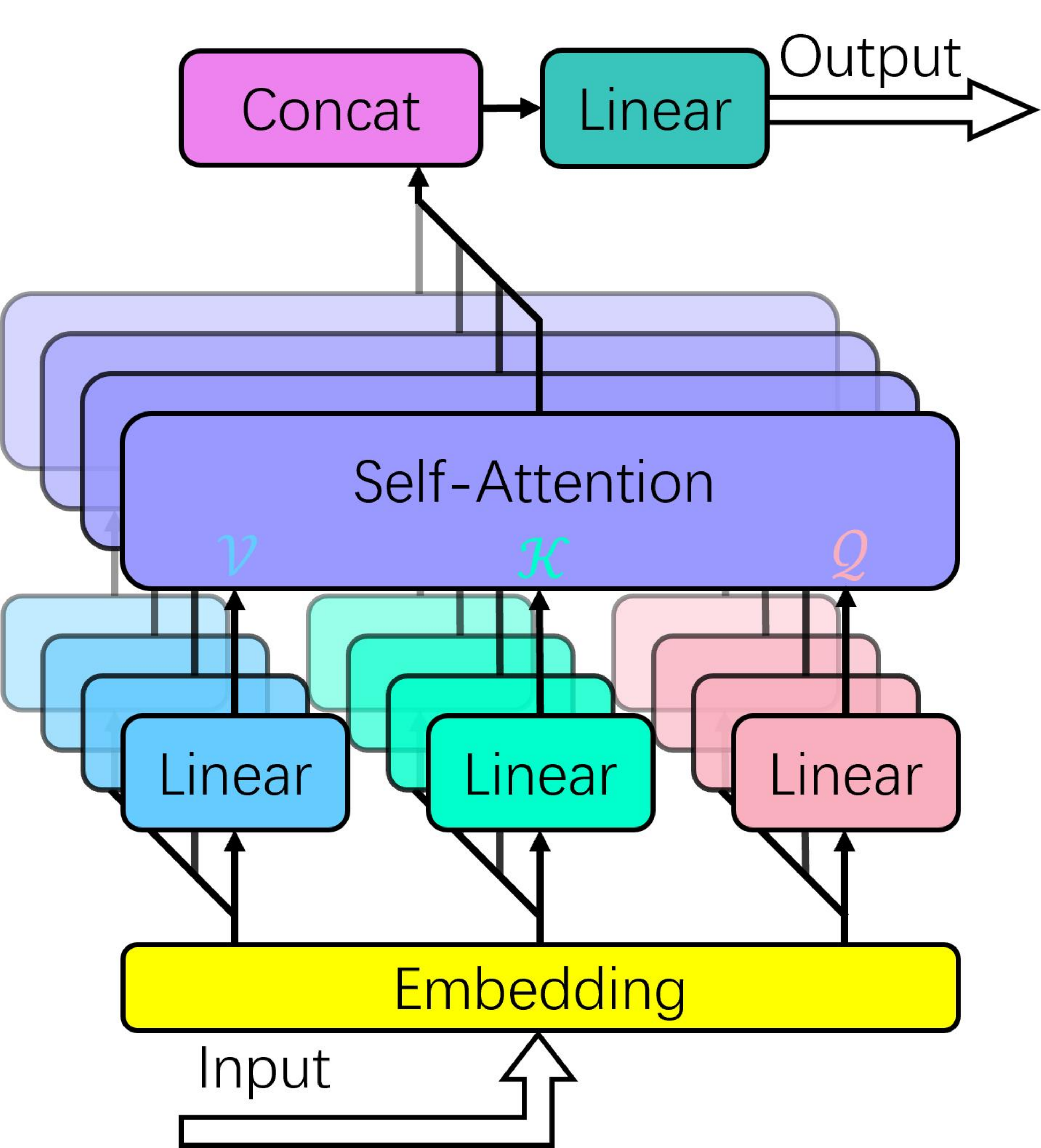}
        \caption{MHSA structure diagram}
        \label{figmhsa}
    \end{subfigure}
    \hfill
    \begin{subfigure}[b]{0.63\textwidth}
        \centering
        \includegraphics[width=\textwidth]{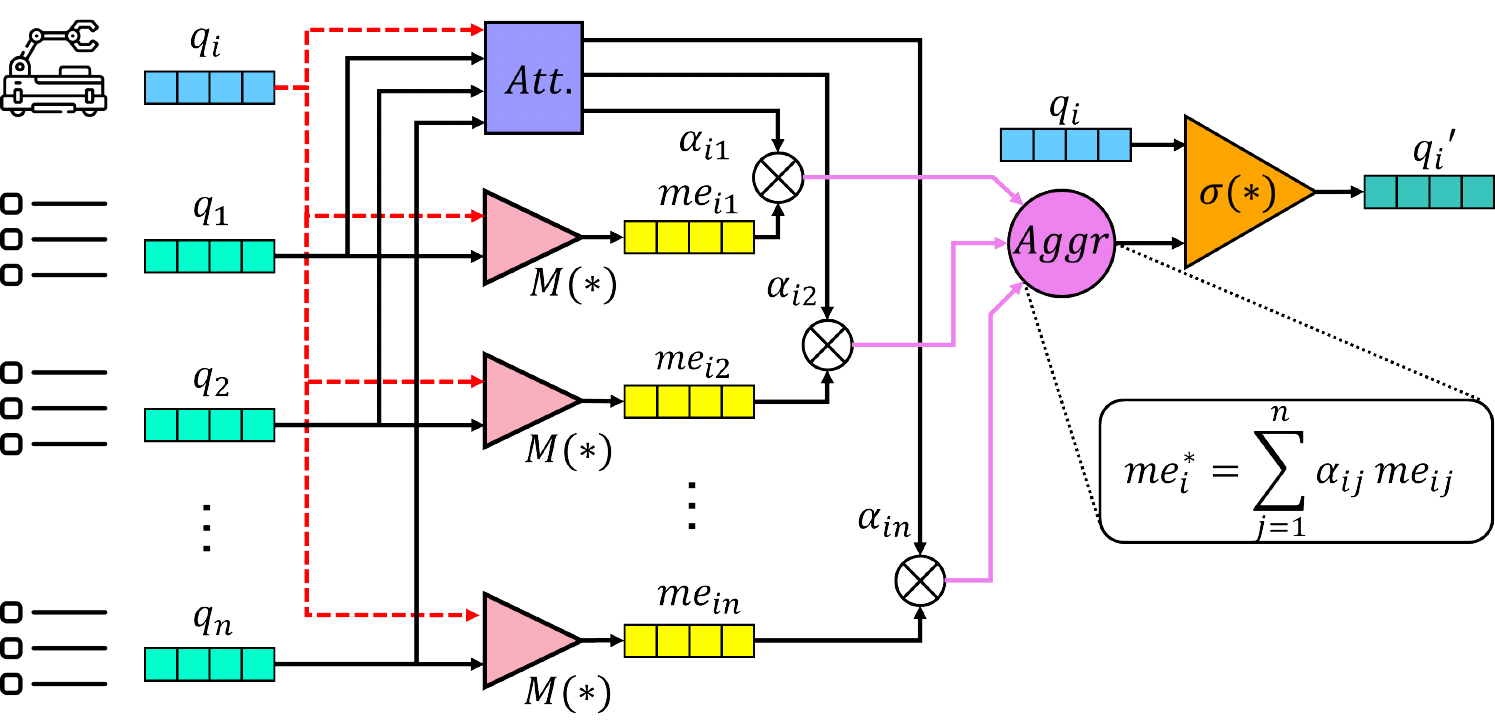}
        \caption{Graph attention structure diagram}
        \label{figgat}
    \end{subfigure}
    \caption{Illustration of the MHSA and graph attention components in the generator.}
    \label{figmg}
\end{figure}

To capture temporal dependencies within inter-task trajectory segments, we employ a MHSA mechanism as part of the representation encoder. For an agent $\lambda_i$, the inter-task trajectory segment \( Traj_{ki} \), together with its temporal ordering, is first mapped into a high-dimensional embedding space. As illustrated in Fig.~\ref{figmhsa}, the embedded representation at each timestep $\iota$ is defined as: $eb_{ki}^{(\iota)} = \sigma \left( W_p [x_{\iota}, y_{\iota}]^\top + b_p \right) + W_{\iota} [1, {\iota}]^\top + b_\iota$, where \( W_p, W_{\iota} \in \mathbb{R}^{d \times 2} \) denote the learnable coordinate embedding matrix and time encoding matrix, respectively, while \( b_p, b_{\iota} \in \mathbb{R}^d \) represent the coordinate embedding bias and time encoding bias, respectively. The parameter \( d \) is the dimension of the embedding layer, and \( \sigma(*) \) is the non-linear activation function. After embedding the trajectory, the embedding matrix is obtained as \( EB_{ki} = [{eb_{ki}^{(1)}, eb_{ki}^{(2)}, \dots,eb_{ki}^{({\iota})} ,\dots, eb_{ki}^{(L)}]}^\top \in \mathbb{R}^{L \times d} \).

Then, the Query, Key and Value required for computing the attention mechanism are defined as: \( \mathcal{Q} = EB_{ki} W_\mathcal{Q} \), \( \mathcal{K} = EB_{ki} W_\mathcal{K} \), \( \mathcal{V} = EB_{ki} W_\mathcal{V} \), 
where \( W_\mathcal{Q}, W_\mathcal{K}, W_\mathcal{V} \in \mathbb{R}^{d \times d_{head}} \) are learnable linear transformation matrices applied to the input embeddings, and \( d_{head} \) is the dimension of a single attention head. For an attention head \( \text{head}_n \), the output is:

\begin{equation}
\label{head}
\text{head}_n = \text{softmax}\left(\frac{\mathcal{Q}_n \mathcal{K}_n^T}{\sqrt{d_{head}}}\right) \mathcal{V}_n .
\end{equation}

Subsequently, the outputs of all \( h \) attention heads are concatenated and fused using a linear layer. The result is the feature matrix 
\begin{equation}
 H_{ki} = \text{Concat}(\text{head}_1, \text{head}_2, \dots, \text{head}_h)\cdot W_O,
\end{equation}
\noindent where \( H_{ki}\in \mathbb{R}^{L \times d} \) and \( W_O \in \mathbb{R}^{(h \cdot d_{\text{head}}) \times d} \) is the final linear transformation matrix for MHSA.

While $H_{ki}$ captures temporal dependencies along inter-task trajectory segments, it does not explicitly encode interactions among agents and tasks that are critical for task allocation decisions. In multi-agent task allocation, an agent’s decision is often influenced not only by its own motion history, but also by shared environmental context, task competition, and cooperative or conflicting behaviors among agents. Such global contextual information cannot be fully represented by temporal features alone and therefore requires an explicit relational modeling mechanism.

The agent–task allocation problem naturally admits a graph-structured representation, in which agents and tasks are modeled as nodes and their interactions are represented as edges. In this structure, different neighboring agents or tasks may contribute unequally to decision making, depending on factors such as spatial proximity, task urgency, and resource contention. GAT is well suited for modeling these relations, as they enable adaptive weighting of neighboring nodes while preserving permutation invariance. Moreover, the inductive bias of GAT aligns well with the dynamic and scalable nature of MATA, allowing the learned representations to generalize across varying numbers of agents and tasks and to adapt to changes in the environment.

As shown in Fig.~\ref{figgat}, for agent \( \lambda_i \) at a given iteration, its feature vector is \( q_i \). The feature vector \( q_i \) and the feature vector \( q_k \) of task \( j_k \) are processed through a message generation function \( M(*) \) to produce the message \( me_{ik} = M(q_i, q_k) \). Simultaneously, \( q_i \) and \( q_k \) are processed through an attention mechanism to generate the attention weight \( \alpha_{ik} \):
\vspace{-2pt}
\begin{equation}
\label{eq_18}
\alpha_{ik} = \frac{\exp \left( \sigma \left( a^T [Wq_i \, || \, Wq_k] \right) \right)}
{\sum_{u \in \mathcal{N}(i)} \exp \left( \sigma \left( a^T [Wq_i \, || \, Wq_{u}] \right) \right)} ,
\end{equation}
where $\Vert$ denotes feature concatenation, $W$ is a learnable linear transformation matrix, $a$ is the attention vector, $\sigma(\cdot)$ denotes a nonlinear activation function, and $\mathcal{N}(i)$ represents the set of neighbors of node $i$.

The messages from neighboring nodes are aggregated using the attention weights to obtain the aggregated message $me_i^* = \sum_{u \in \mathcal{N}(i)} \alpha_{iu} \, me_{iu}$. The node feature of agent $\lambda_i$ is then updated as
\begin{equation}
\label{eq_19}
q_i' = \sigma \left( W_q q_i + me_i^* \right),
\end{equation}
where $W_q$ is a learnable weight matrix.

After temporal and relational representations are obtained, the outputs of MHSA and GAT are jointly used to generate reward adaptation coefficients. Specifically, the temporal feature matrix $H_{ki}$ is first summarized along the trajectory dimension to obtain a compact representation of inter-task behavior:\(\bar{h}_{ki} = \frac{1}{L} \sum_{v=1}^{L} H_{ki}^{(v)}\). The relational feature $q_i'$, produced by the GAT layer, encodes global interaction information between agent $\lambda_i$ and surrounding tasks. These two complementary representations are then fused by concatenation to form a unified feature vector \(f_{ki} = [\bar{h}_{ki} \, \Vert \, q_i']\).

Based on the fused representation $f_{ki}$, a lightweight projection is applied to produce the reward adaptation coefficients $\alpha$ and $\beta$:
\begin{equation}
[\alpha, \beta]^\top = W_r f_{ki} + b_r
\end{equation}

\noindent where $W_r$ and $b_r$ are learnable parameters. This design allows temporal behavior patterns and relational context to jointly influence reward adaptation while maintaining a compact and interpretable parameterization. In practice, this projection is implemented using a small multilayer perceptron (MLP), which provides sufficient expressive capacity while maintaining training stability. Although the representation is constructed at the inter-task level, the reward adaptation operates at a global scale via shared coefficients.

\renewcommand{\algorithmicrequire}{\textbf{INPUT:}}

\subsection{Algorithm Description}

This section summarizes the operational details of the proposed framework. Algorithm  \ref{algorithm_irl} details the adversarial training procedure of the proposed reward inference module. The generation phase in lines 3 to 10 focuses on inferring adaptive reward coefficients from trajectory segments, while the adversarial phase in lines 11 to 18 updates the generator and discriminator to align the inferred rewards with expert preferences. Algorithm \ref{algorithm_marl} describes the process of multi-agent training that integrates IRL. Lines 7 to 13 describe the interaction loop, where agents select actions, execute joint behavior, and collect states and rewards. The IRL module is invoked in line 14 to refine the reward signals based on expert demonstrations. Lines 15 to 20 correspond to the learning phase, where experiences are stored and agents update their policies once enough samples are accumulated. Through repeated interaction, reward refinement, and policy updates, the multi-agent system gradually improves coordination performance.

\begin{algorithm}
\normalsize
\caption{Train IRL Part}
\label{algorithm_irl}
\begin{algorithmic}[1]
\REQUIRE number of heads $H$, model dimensionality $d$, learning rate of discriminator $\eta^{d}$
\STATE Initialize the generator with parameters $H$, $d$
\STATE Initialize the discriminator with parameters $\eta^{d}$
\FOR{$i = 1$ to $N$}
    \IF{Agent $\lambda_{i}$ finishes task $j_{k}$}
        \STATE Generate coefficients $\alpha$ and $\beta$ by generator
        \STATE Obtain the revised reward  $r_{i,t}^{IRL} = \alpha \cdot r_{i,t} + \beta$
    \ELSE
        \STATE Set original reward as the revised reward $r_{i,t}^{IRL} = r_{i,t}$
    \ENDIF
\ENDFOR
\FOR{$i = 1$ to $N$}
    \IF{Agent $\lambda_{i}$ finish task $j_{k}$}
        \STATE Obtain the trajectory $Traj_{ki}$ and its probability of being identified as an expert trajectory $P_{ki}$
        \STATE Update the generator by minimizing the loss defined in (\ref{loss_gen})
        \STATE Update the discriminator by minimizing the loss defined in (\ref{loss_dis})
    \ENDIF
\ENDFOR
\end{algorithmic}
\end{algorithm}

\begin{algorithm}[t!]
\normalsize
\caption{Train Multi-Agent System}
\label{algorithm_marl}
\begin{algorithmic}[1]
\REQUIRE number of training episodes $N_{max}$, number of steps in an episode $t_{max}$, learning rate of agents $\eta$, discount factor $\gamma$, batch size $\mathcal{B}$, replay buffer $\mathcal{D}$
\FOR{$i = 1$ to $N$}
    \STATE Initialize the agent with parameters $\eta$
\ENDFOR
\FOR{$episode = 1$ to $N_{max}$}
    \STATE Reset the environment
    \FOR{$t = 1$ to $t_{max}$}
        \FOR{$i = 1$ to $N$}
            \STATE Select an action $a_{i,t}$ for agent $\lambda_{i}$
        \ENDFOR
        \STATE Obtain the state $s_{t}$
        \STATE Execute the joint action $\mathbf{a}_{t} = (a_{1,t}, a_{2,t}, a_{3,t}, \dots, a_{n,t})$
        \STATE Obtain the new state $s_{t+1}$
        \STATE Obtain the  reward $\mathbf{r}_{t} = (r_{1,t}, r_{2,t}, r_{3,t}, \dots, r_{n,t})$
        \STATE \textbf{(IRL)} Call Algorithm~\ref{algorithm_irl} to obtain revised rewards $\{r_{i,t}^{IRL}\}_{i=1}^{N}$ and update the generator/discriminator if applicable
        \STATE Store $(s_{t}, \mathbf{a}_{t}, \mathbf{r}_{t}, s_{t+1})$ in replay buffer $\mathcal{D}$
        \IF{the batch size of data stored in $\mathcal{D}$ is greater than $\mathcal{B}$}
            \FOR{$i = 1$ to $N$}
                \STATE Update the agent $\lambda_{i}$ by minimizing the loss defined in (\ref{loss_Q}) and (\ref{loss_pi})
            \ENDFOR
        \ENDIF
        \STATE Update the state $s_{t}=s_{t+1}$
    \ENDFOR
\ENDFOR
\end{algorithmic}
\end{algorithm}

\section{Experiments and Results}
\label{sec:results_discussion}
In order to verify the effectiveness and efficiency of the proposed approach, experiments are conducted in the open-source multi-agent environment, which consists of agents and tasks inhabiting a two-dimensional world with continuous space and discrete time. A comparison is conducted between the proposed method and two widely used MARL algorithms, namely MASAC \cite{gupta2019probabilistic} and MAPPO \cite{yu2022MAPPO}.

\subsection{Experiment Setup}
The experiments are conducted in an open-source MARL environment developed by OpenAI. The environment consists of multiple agents interacting in a two-dimensional continuous space with discrete timesteps. The detailed environment settings are summarized in Table \ref{tab:env_set}.

The hyperparameters of the proposed algorithm, including the learning rate, batch size and exploration strategy, are listed in Table \ref{tab:hypar}. The experiments are performed on a system equipped with an AMD Ryzen 9 5950X CPU, 32GB RAM, and an Nvidia RTX 3080Ti GPU.

To provide expert demonstrations for the IRL framework, expert trajectories were generated using representative convex optimization-based task allocation methods \cite{shorinwa2023Distributed} under a fully static environment. Although these algorithms are not directly applicable to large-scale or highly dynamic scenarios due to their computational complexity and lack of online adaptability, they can still produce structured and near-optimal trajectories in static settings. Importantly, these trajectories are not used as final solutions but as high-quality priors for reward inference. Within the IRL framework, the inferred reward functions generalize beyond the static demonstrations and enable the policy to adapt to dynamic and uncertain environments, thereby mitigating the limitations of the original optimization methods.

\begin{figure}
    \centering

    \begin{subfigure}[t]{0.4\textwidth}
        \centering
        \includegraphics[width=\textwidth]{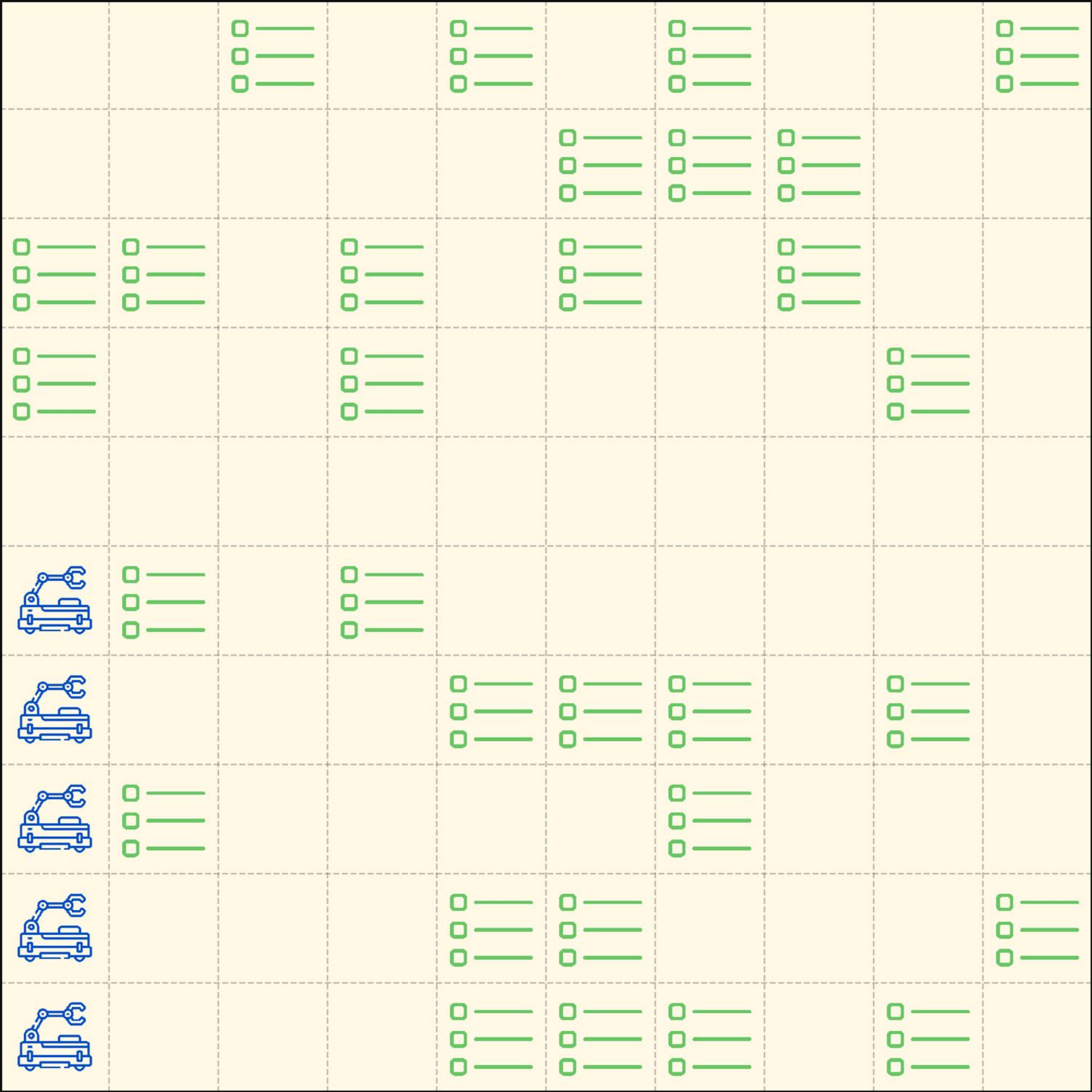}
        \caption{Initial spatial distribution of agents and tasks}
        \label{init}
    \end{subfigure}%
    \hfill
    \begin{subfigure}[t]{0.4\textwidth}
        \centering
        \includegraphics[width=\textwidth]{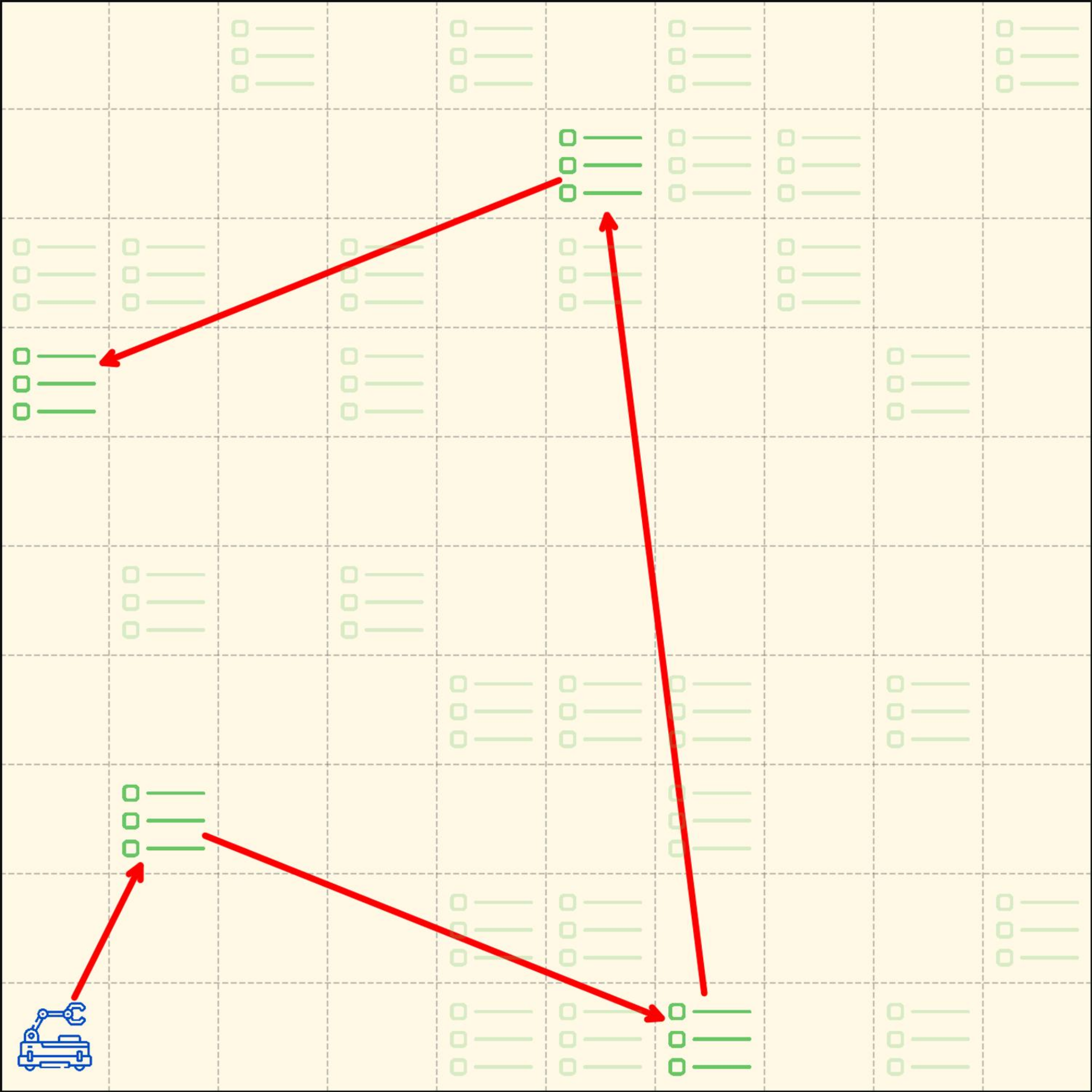}
        \caption{Illustrative example of an agent’s movement}
        \label{run}
    \end{subfigure}%
    \hfill
    \begin{subfigure}[t]{0.125\textwidth}
        \centering
        \includegraphics[width=\textwidth]{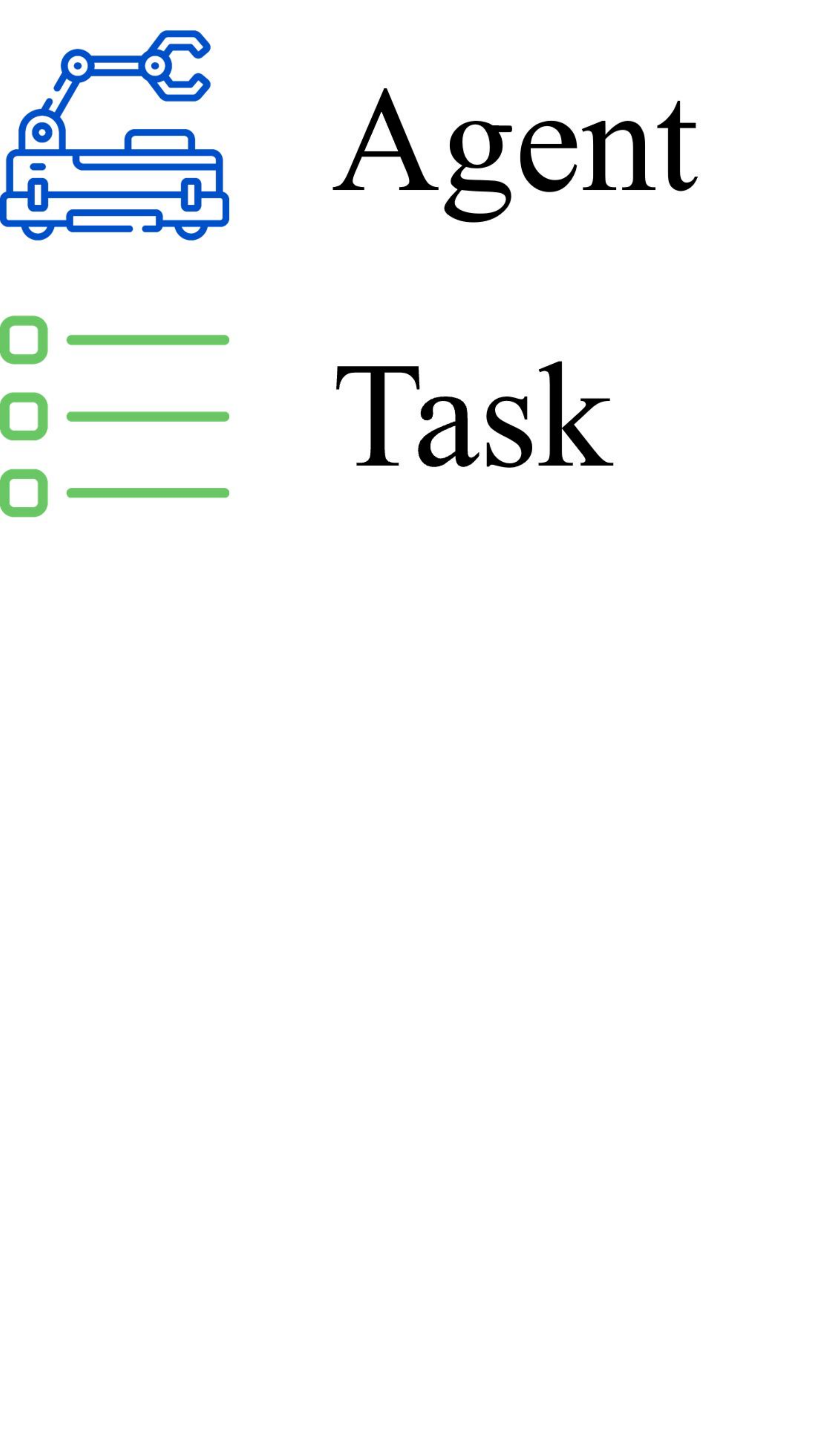}
    \end{subfigure}

    \caption{Illustrations of the experimental environment.}
\end{figure}

Fig.~\ref{run} provides an illustrative example of the motion of a single agent as it sequentially completes several tasks. The red arrows indicate the visiting order of the selected tasks and the corresponding trajectory segments between them.

These visualization graphs help readers understand the geometric layout and task distribution in our experimental setting, and they serve purely as schematic illustrations of the environment configuration rather than depictions of any actual algorithm run. The numbers of agents and tasks shown in these figures do not reflect the scales used in the experiments reported in Section \ref{pa}, and the motion example does not represent the true trajectories produced by our method or by any baseline. Instead, these simplified illustrations are intended only to provide qualitative intuition regarding the spatial arrangement of entities and the general form of agent–task interactions. 

\begin{table}[width=.8\linewidth,cols=2,pos=t]
\caption{Environment Settings}
\label{tab:env_set}

\begin{tabular*}{\tblwidth}{@{} L C @{}}
\toprule
\textbf{Parameter} & \textbf{Value} \\
\midrule
Number of Agents & 5 / 10 / 20 \\
Number of Tasks & 20 / 40 / 60 \\
Environment Size & 20 $\times$ 20 \\
Speed of Agents & 5 \\
Timesteps Required for each Agent to Complete one Task & 1 \\
Reward for Completing a Task & 7.5 \\
Time Consumption Penalty & 0.5 \\
Energy Consumption Penalty & 1.5 \\
\bottomrule
\end{tabular*}
\end{table}

\begin{table}[width=.8\linewidth,cols=2,pos=t]
\caption{Hyperparameters}
\label{tab:hypar}

\begin{tabular*}{\tblwidth}{@{} L C @{}}
\toprule
\textbf{Parameter} & \textbf{Value} \\
\midrule
Number of Training Episodes $N_{\max}$ & 2000 \\
Number of Steps in an Episode $t_{\max}$ & 300 \\
Learning Rate of Actors $\eta^{a}$ & $5\times10^{-5}$ \\
Learning Rate of Critics $\eta^{c}$ & $10^{-5}$ \\
Discount Factor $\gamma$ & 0.95 \\
Batch Size $\mathcal{B}$ & 2048 \\
Replay Buffer $\mathcal{D}$ & $10^{6}$ \\
Number of Heads $H$ & 16 \\
Model Dimensionality $d_H$ of MHSA Mechanism & 256 \\
Learning Rate of Generator $\eta^{g}$ & $10^{-5}$ \\
Learning Rate of Discriminator $\eta^{d}$ & $2\times10^{-5}$ \\
\bottomrule
\end{tabular*}
\end{table}

Because multi-agent trajectories often overlap heavily and become visually indistinguishable when plotted together, such figures cannot meaningfully convey policy differences or performance characteristics. Consequently, they should not be interpreted as evidence of algorithmic behavior, nor as representative samples of the dynamic processes occurring in our simulations. All experimental evaluations and method comparisons rely strictly on quantitative performance metrics—including cumulative reward, timestep consumption, and total travel distance—which are reported and analyzed in Section \ref{pa}. Note that while our agents train on the adaptive reward, all evaluation metrics reported in this section are based on the original environmental reward to ensure a consistent benchmark against baselines.

\subsection{Performance Analysis}
\label{pa}

\begin{figure*}
\centering
\subfloat[5 agents with 20 tasks]{\includegraphics[width=0.33\textwidth]{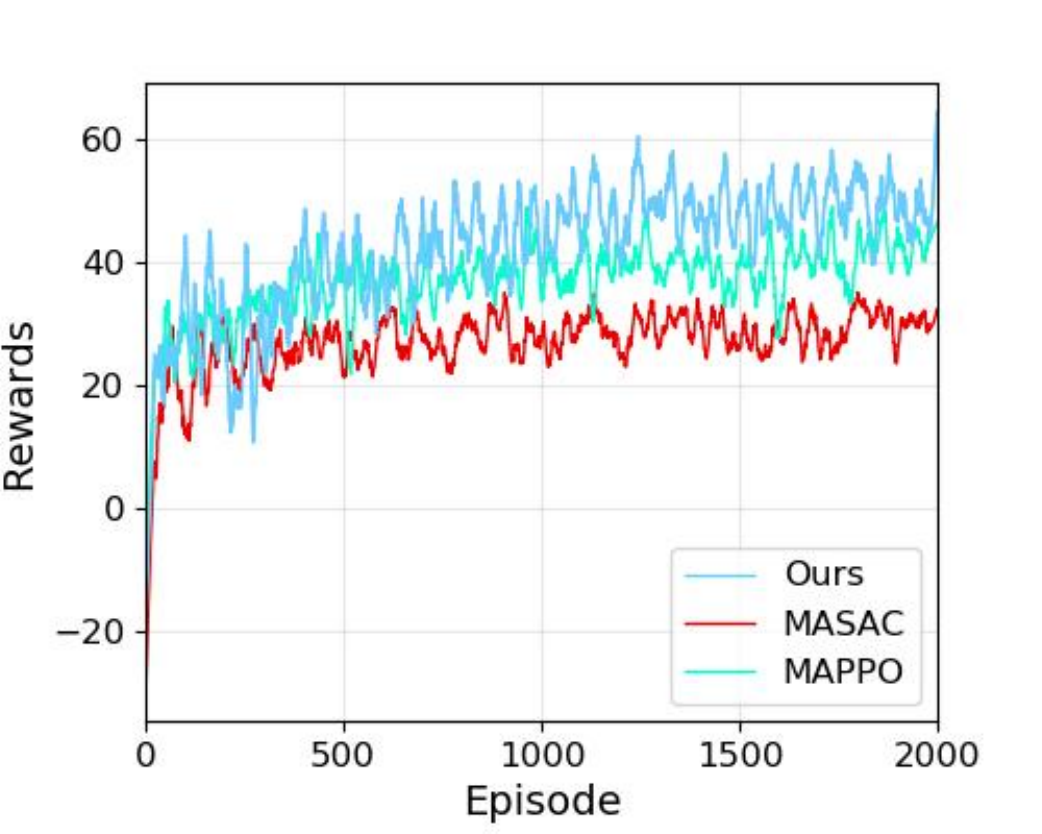}\label{r1}}
\hfil
\subfloat[5 agents with 40 tasks]{\includegraphics[width=0.33\textwidth]{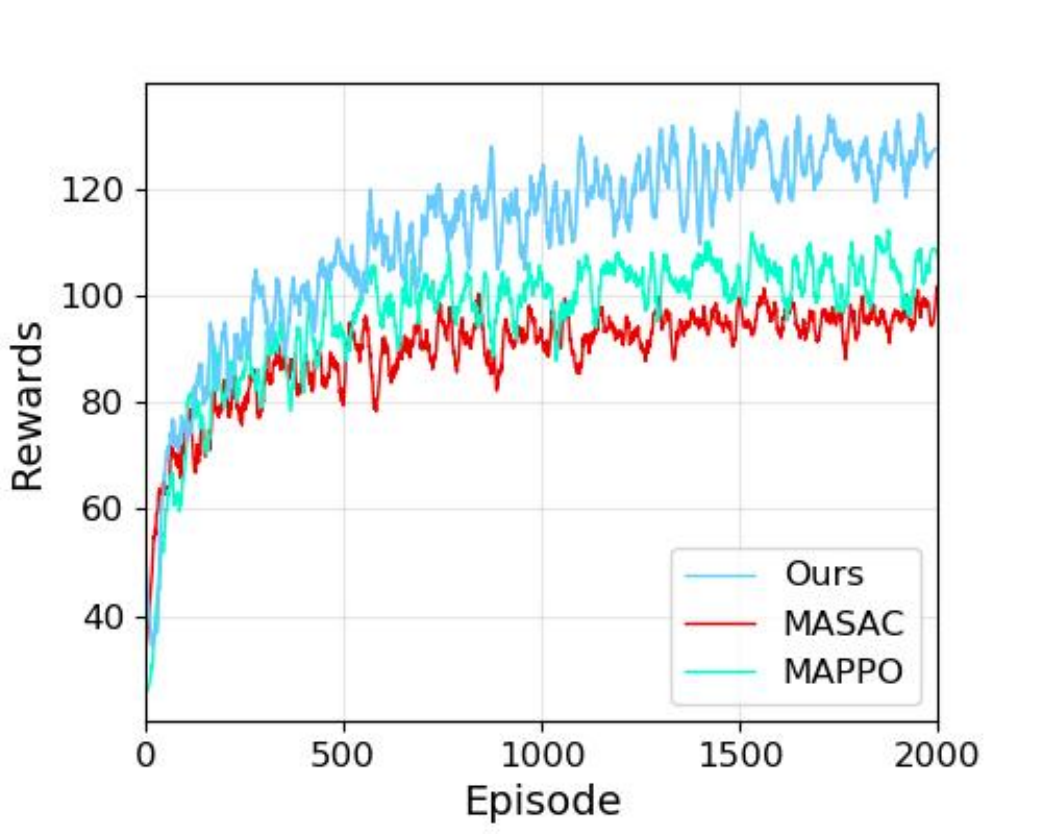}\label{r2}}
\hfil
\subfloat[5 agents with 60 tasks]{\includegraphics[width=0.33\textwidth]{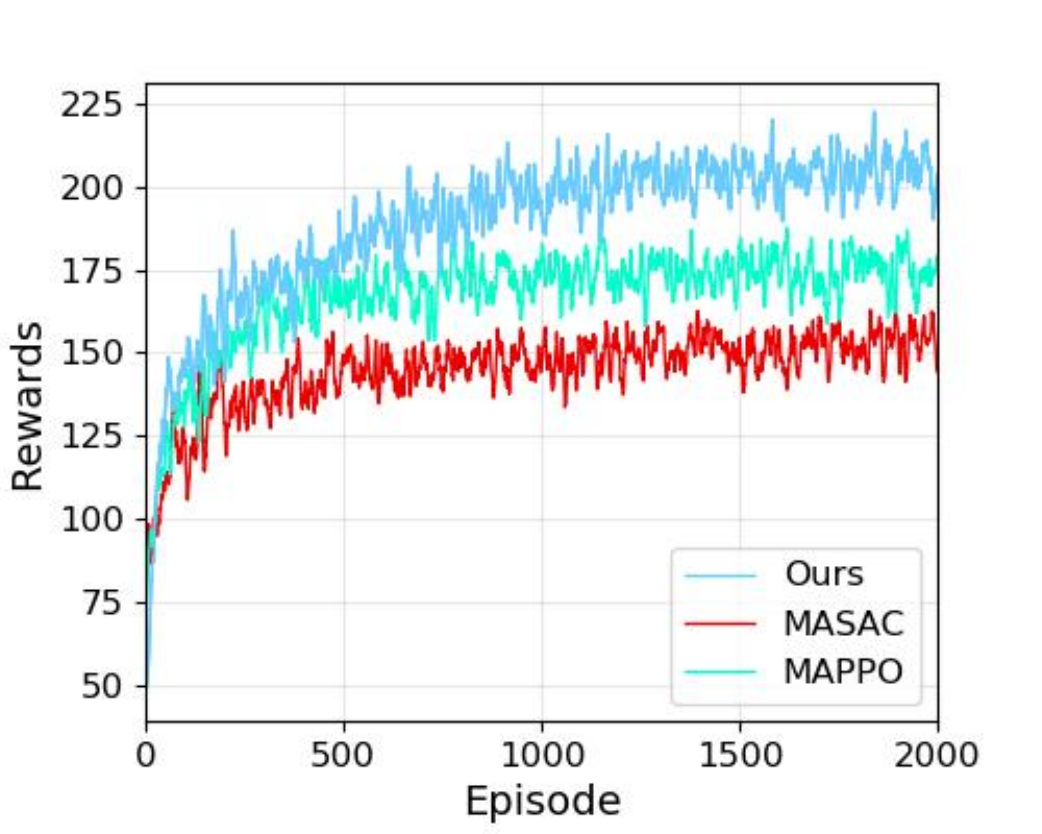}\label{r3}}
\\
\subfloat[10 agents with 20 tasks]{\includegraphics[width=0.33\textwidth]{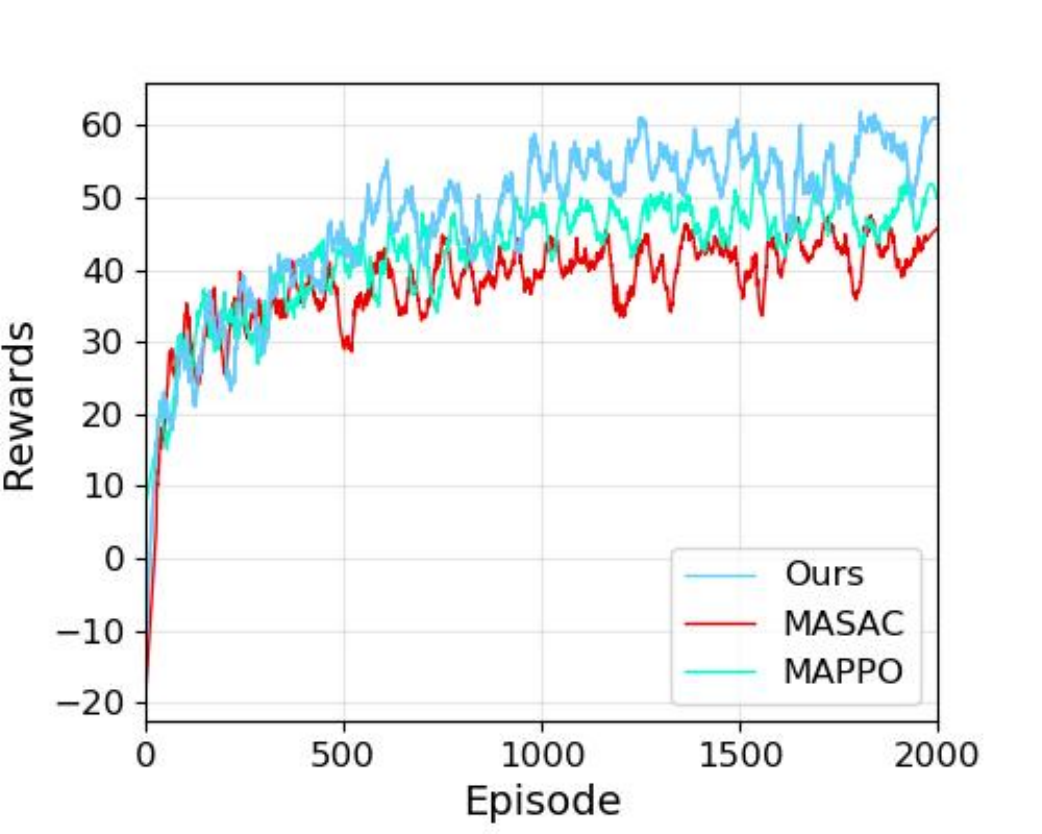}\label{r4}}
\hfil
\subfloat[10 agents with 40 tasks]{\includegraphics[width=0.33\textwidth]{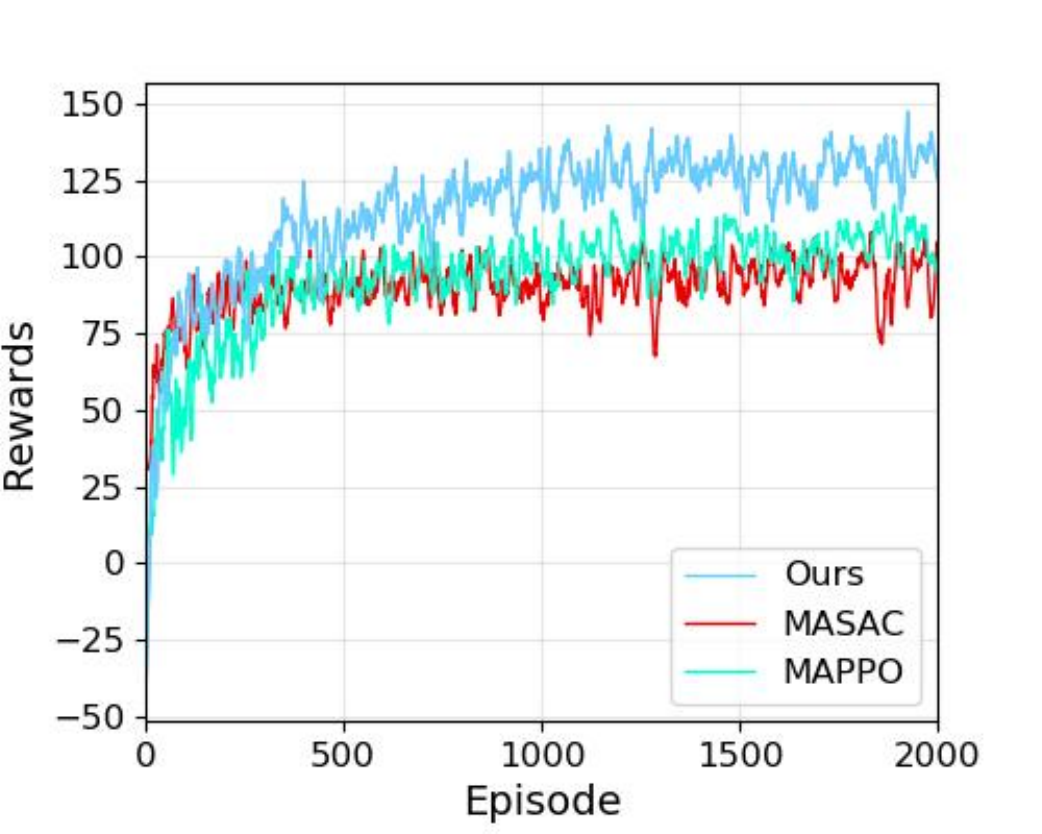}\label{r5}}
\hfil
\subfloat[10 agents with 60 tasks]{\includegraphics[width=0.33\textwidth]{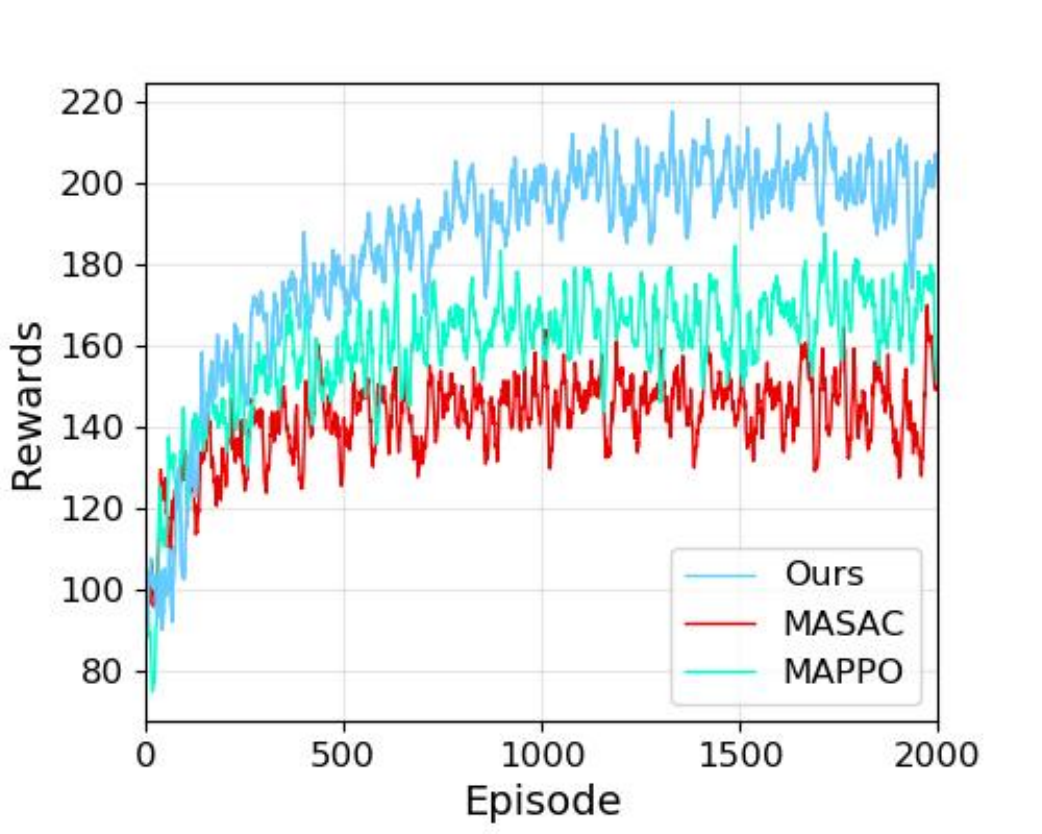}\label{r6}}
\\
\subfloat[20 agents with 20 tasks]{\includegraphics[width=0.33\textwidth]{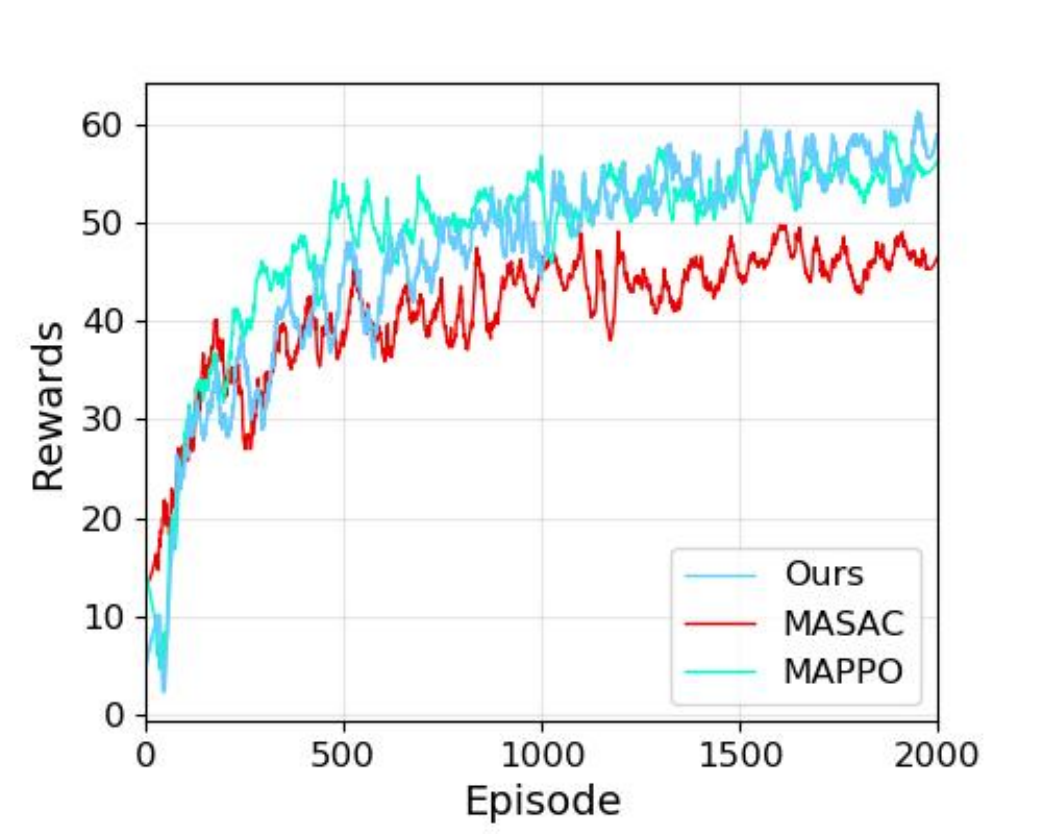}\label{r7}}
\hfil
\subfloat[20 agents with 40 tasks]{\includegraphics[width=0.33\textwidth]{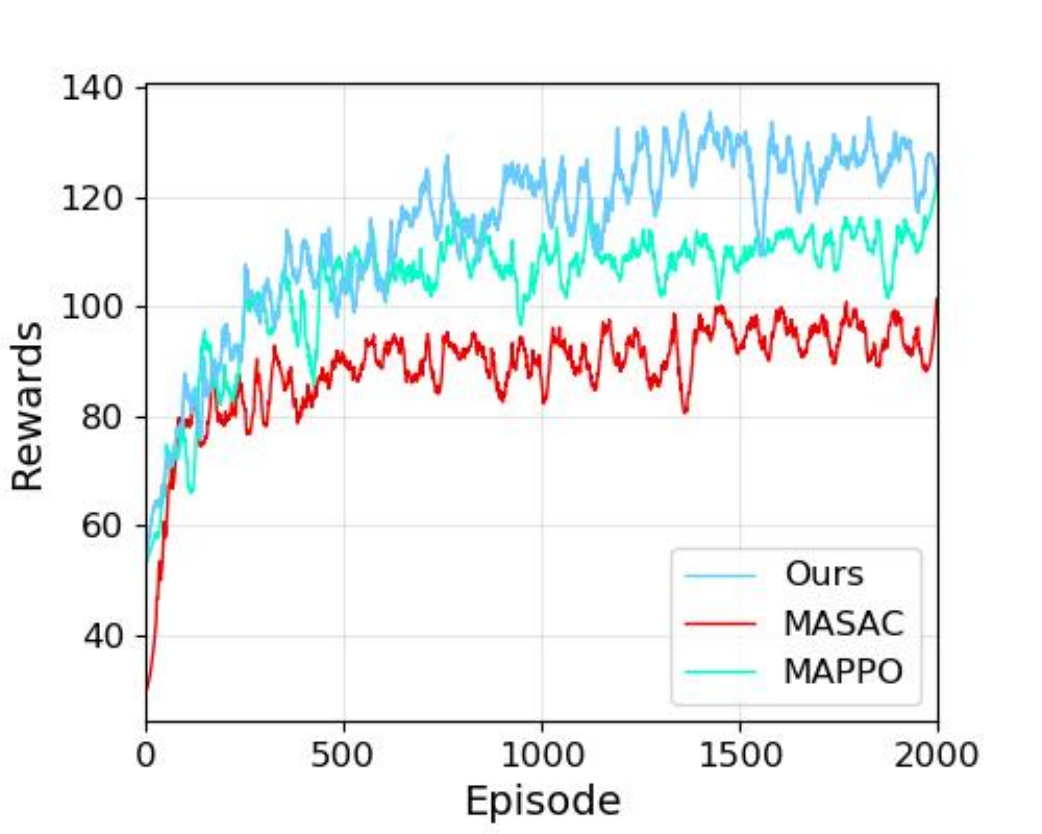}\label{r8}}
\hfil
\subfloat[20 agents with 60 tasks]{\includegraphics[width=0.33\textwidth]{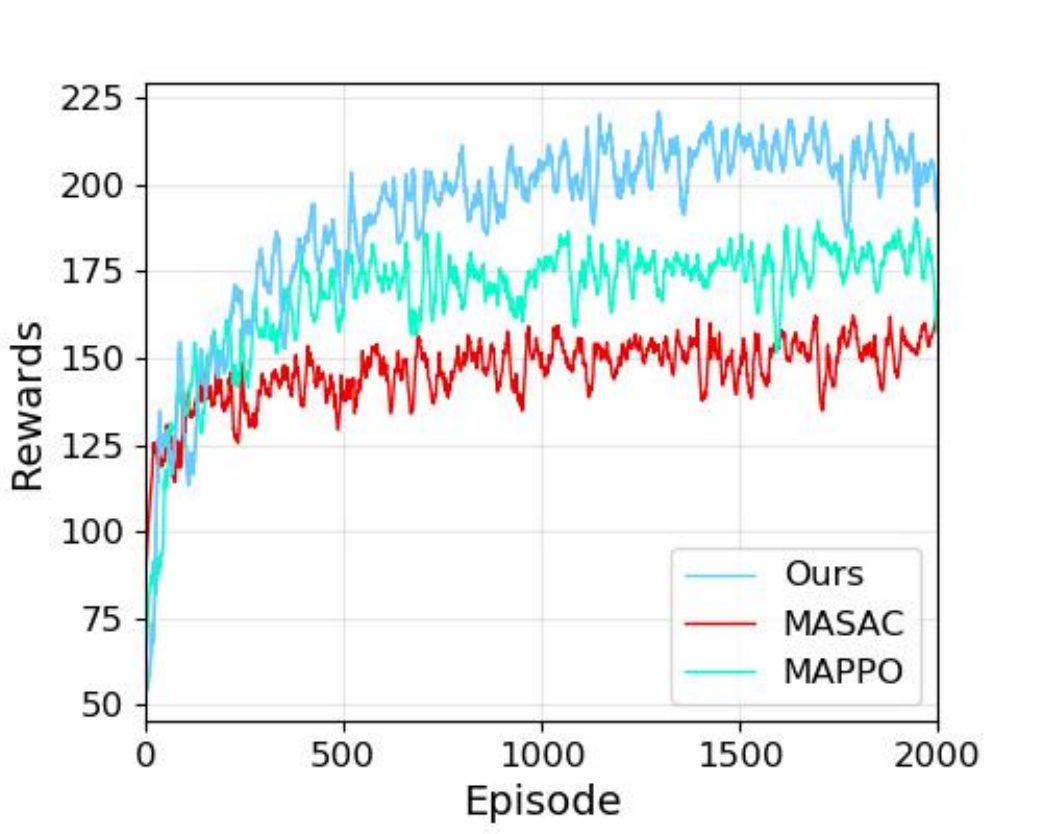}\label{r9}}

\caption{Performance of cumulative rewards in different scenarios.}
\label{fig_reward}
\end{figure*}

In this subsection, experiments are conducted using the aforementioned environment settings and hyperparameters. The performance of the proposed method is assessed based on three key metrics: cumulative reward, time-step consumption, and total travel distance. Cumulative Reward reflects the overall quality of the learned policy. A higher cumulative reward indicates that the allocation strategy achieves better trade-offs between energy efficiency and task completion, aligning with the optimization objective defined in Eq. \ref{eq_4a}. Timestep consumption measures the number of simulation steps required to complete all tasks. Lower values indicate faster task coordination and higher system throughput, which are essential for time-sensitive applications. Total travel distance quantifies the movement cost of all agents during task execution. Shorter total distance not only reduces energy expenditure but also improves spatial efficiency, making it a critical indicator for sustainable multi-agent operations.

Fig.~\ref{fig_reward} illustrates the performance of cumulative reward across varying agent-task scales. Under the heavy-load configuration with 5 agents and 60 tasks, a clear separation emerges after the early episodes. The proposed method attains higher steady-state rewards and maintains lower variance than MASAC and MAPPO, while convergence occurs more rapidly. Such behavior is consistent with the objective in \ref{eq_4a}, where completion time and energy usage must be jointly minimized. The advantage can be attributed to the IRL-based reward revision, which produces signals more closely aligned with implicit operational goals. Furthermore, the attention modules embedded in the generator capture both temporal dependencies and structural interactions, enabling globally coherent task allocations under congestion. These properties together result in superior cumulative returns.  

\begin{figure*}
\centering
\subfloat[5 agents with 20 tasks]{\includegraphics[width=0.33\textwidth]{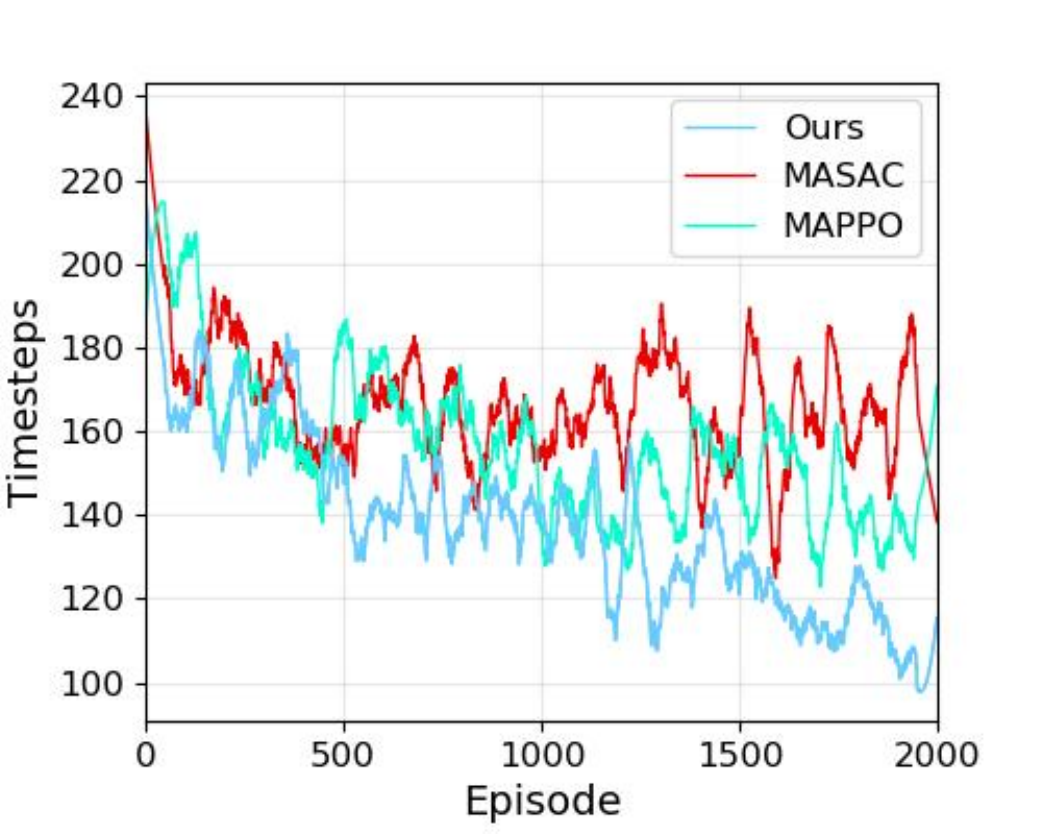}\label{TIMESTEPr1}}
\hfil
\subfloat[5 agents with 40 tasks]{\includegraphics[width=0.33\textwidth]{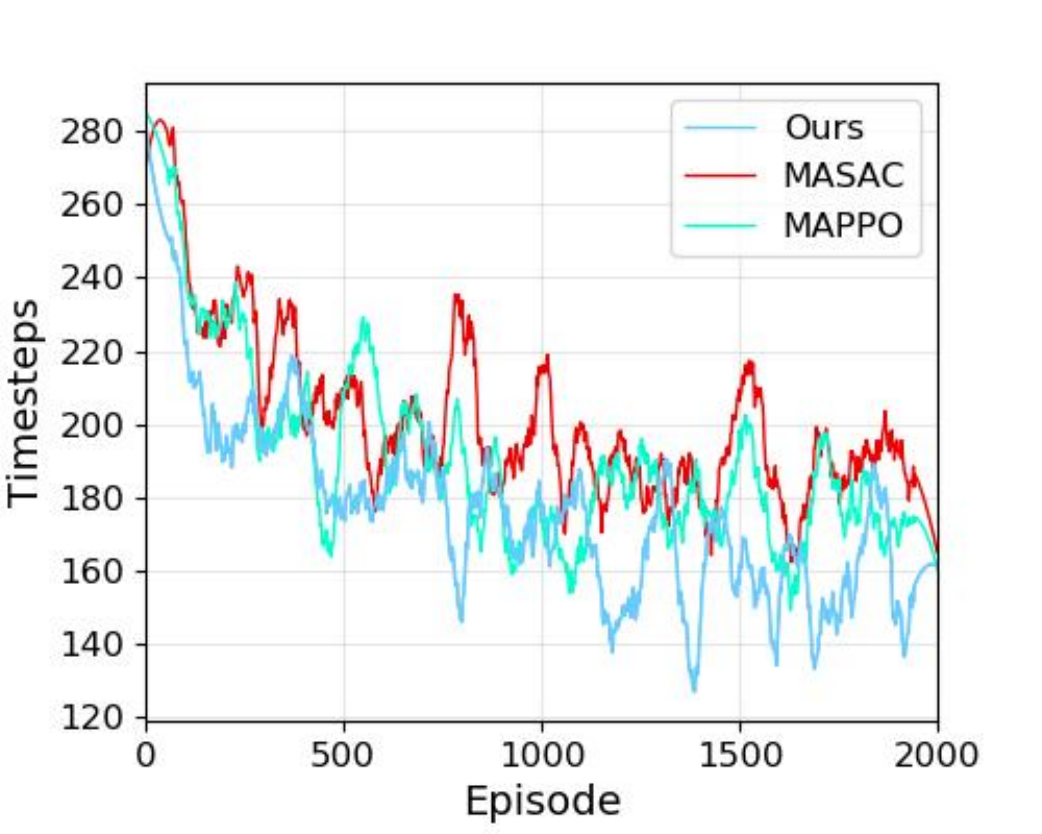}\label{TIMESTEPr2}}
\hfil
\subfloat[5 agents with 60 tasks]{\includegraphics[width=0.33\textwidth]{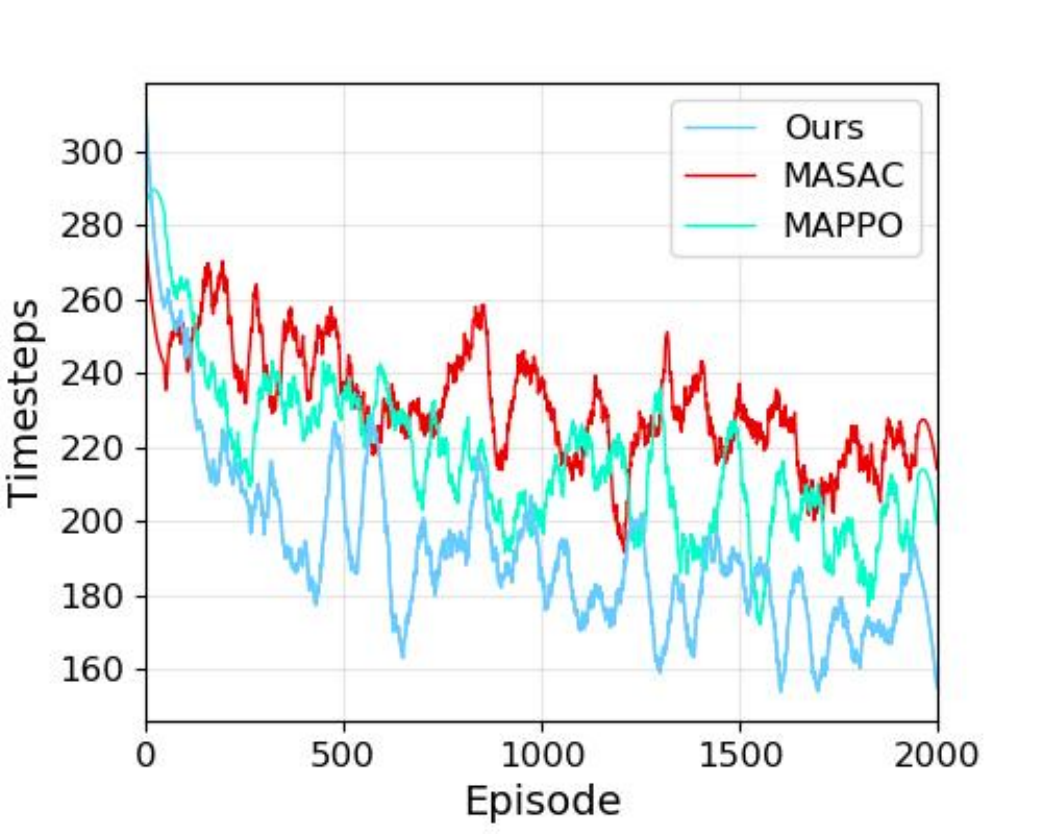}\label{TIMESTEPr3}}
\\
\subfloat[10 agents with 20 tasks]{\includegraphics[width=0.33\textwidth]{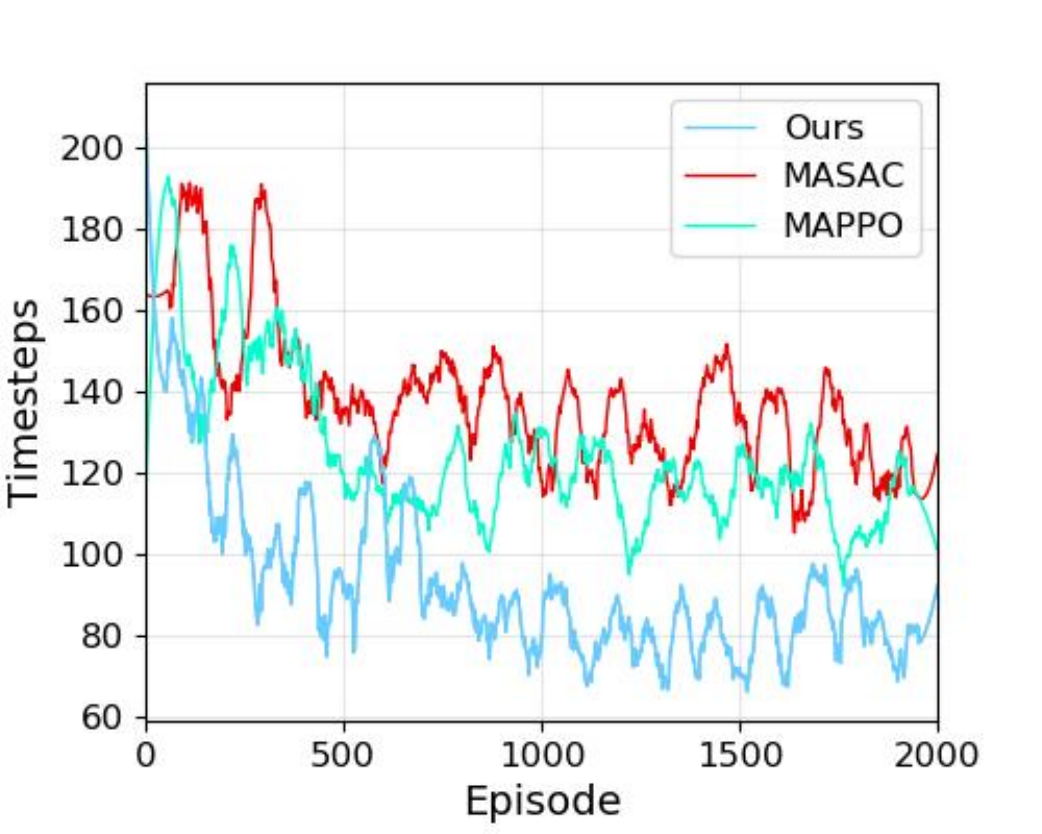}\label{TIMESTEPr4}}
\hfil
\subfloat[10 agents with 40 tasks]{\includegraphics[width=0.33\textwidth]{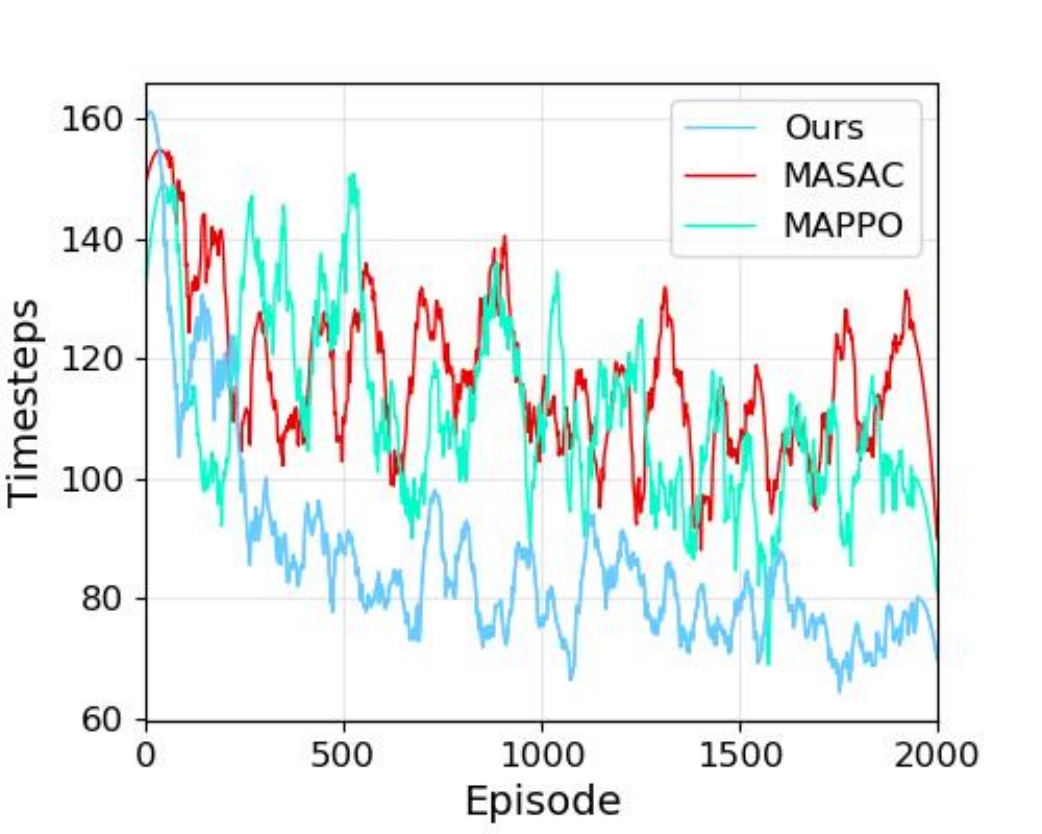}\label{TIMESTEPr5}}
\hfil
\subfloat[10 agents with 60 tasks]{\includegraphics[width=0.33\textwidth]{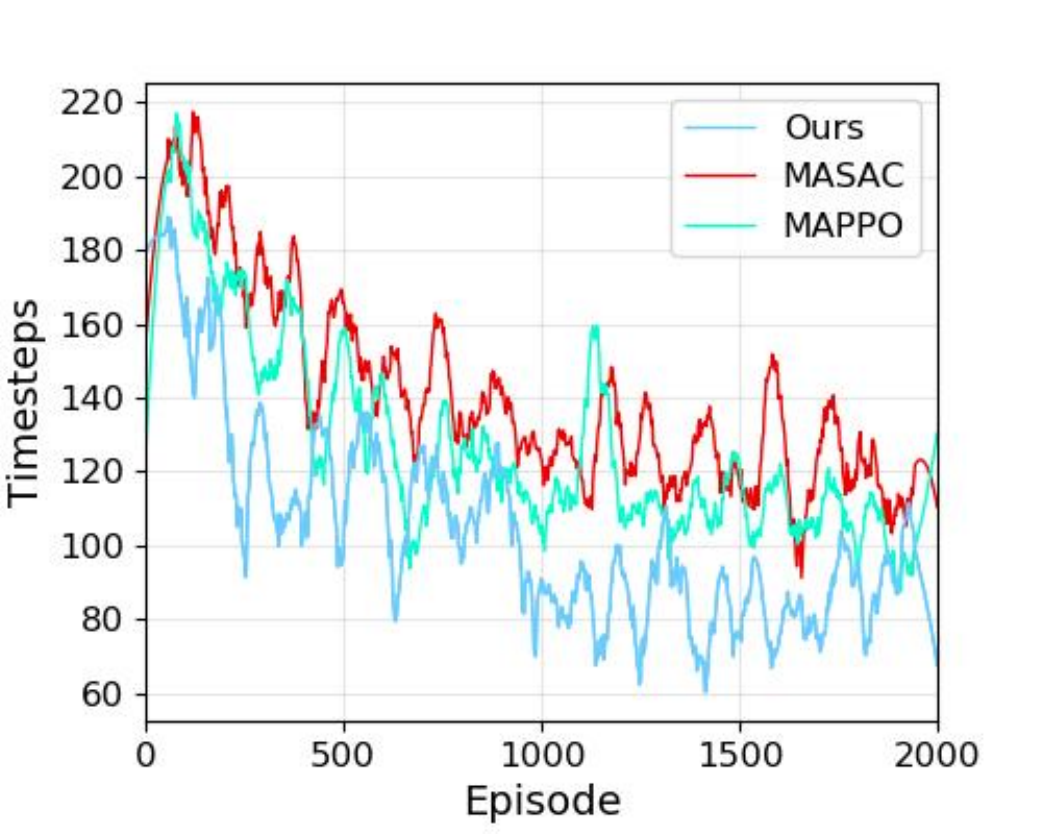}\label{TIMESTEPr6}}
\\
\subfloat[20 agents with 20 tasks]{\includegraphics[width=0.33\textwidth]{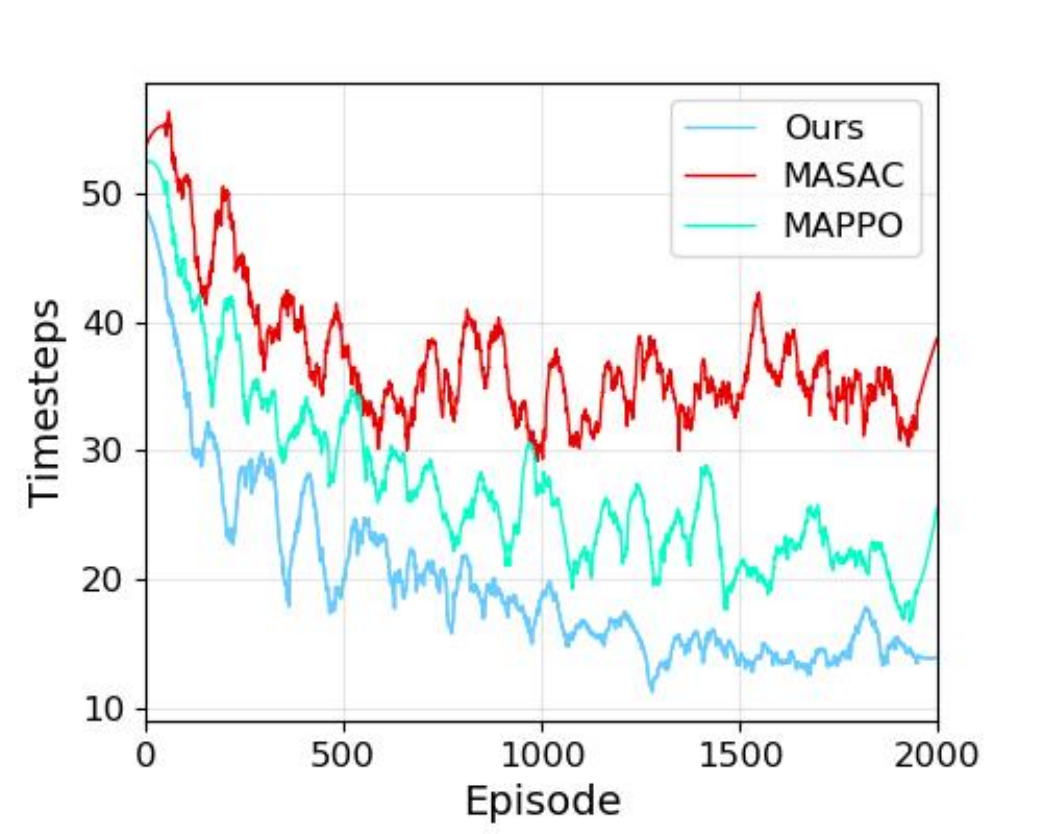}\label{TIMESTEPr7}}
\hfil
\subfloat[20 agents with 40 tasks]{\includegraphics[width=0.33\textwidth]{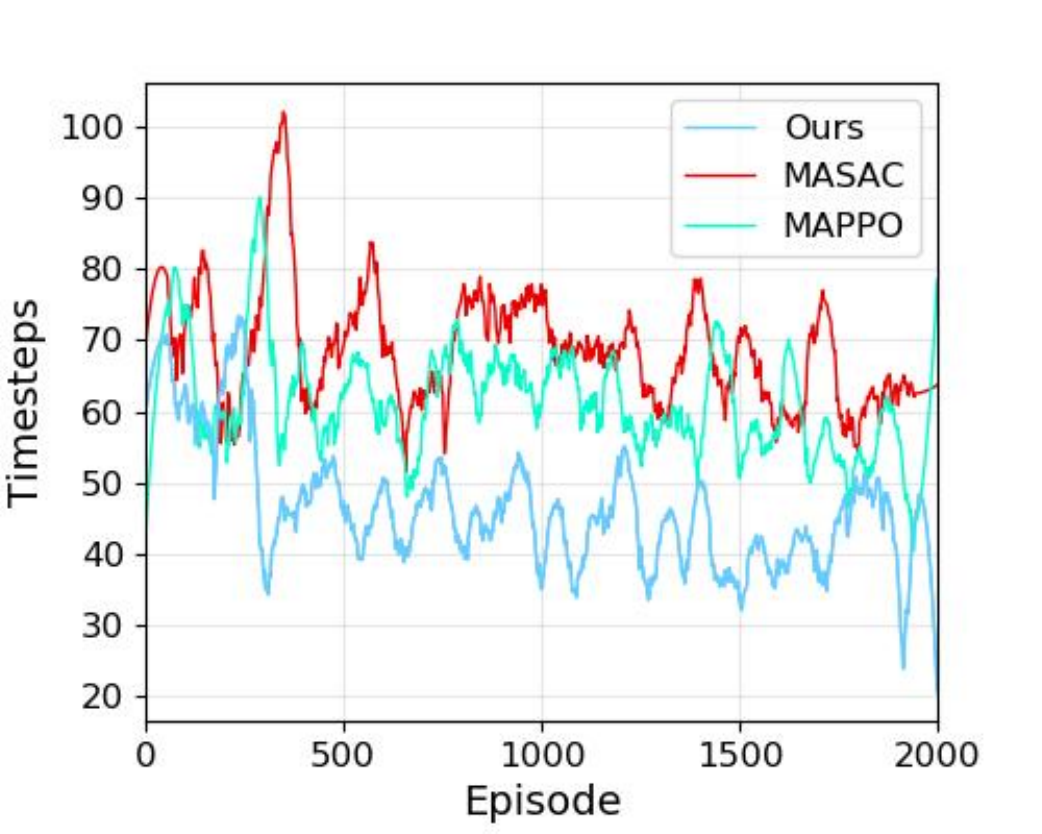}\label{TIMESTEPr8}}
\hfil
\subfloat[20 agents with 60 tasks]{\includegraphics[width=0.33\textwidth]{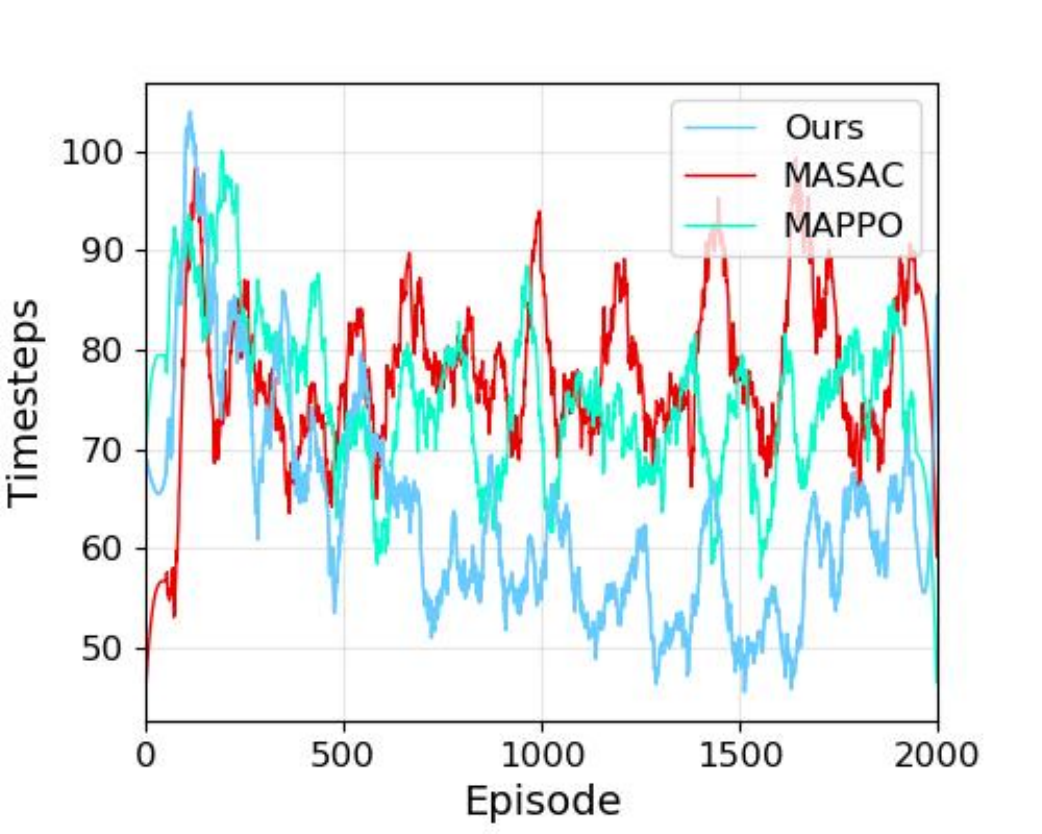}\label{TIMESTEPr9}}

\caption{Performance of timestep consumption in different scenarios.}
\label{fig_TIMESTEP}
\end{figure*}

A similar trend is evident in the scenario with 20 agents and 60 tasks, where scalability becomes a key challenge. Despite the enlarged joint action space and the denser interaction graph, the proposed method continues to achieve higher final rewards and exhibits improved stability compared to the baselines. This outcome indicates that the reward inference mechanism effectively encodes the trade-off between time and energy, while the combination of multi-head self-attention and graph attention provides richer trajectory representations. Within the adversarial imitation framework, these components enhance sample efficiency and stabilize training.

\begin{figure*}
\centering
\subfloat[5 agents with 20 tasks]{\includegraphics[width=0.33\textwidth]{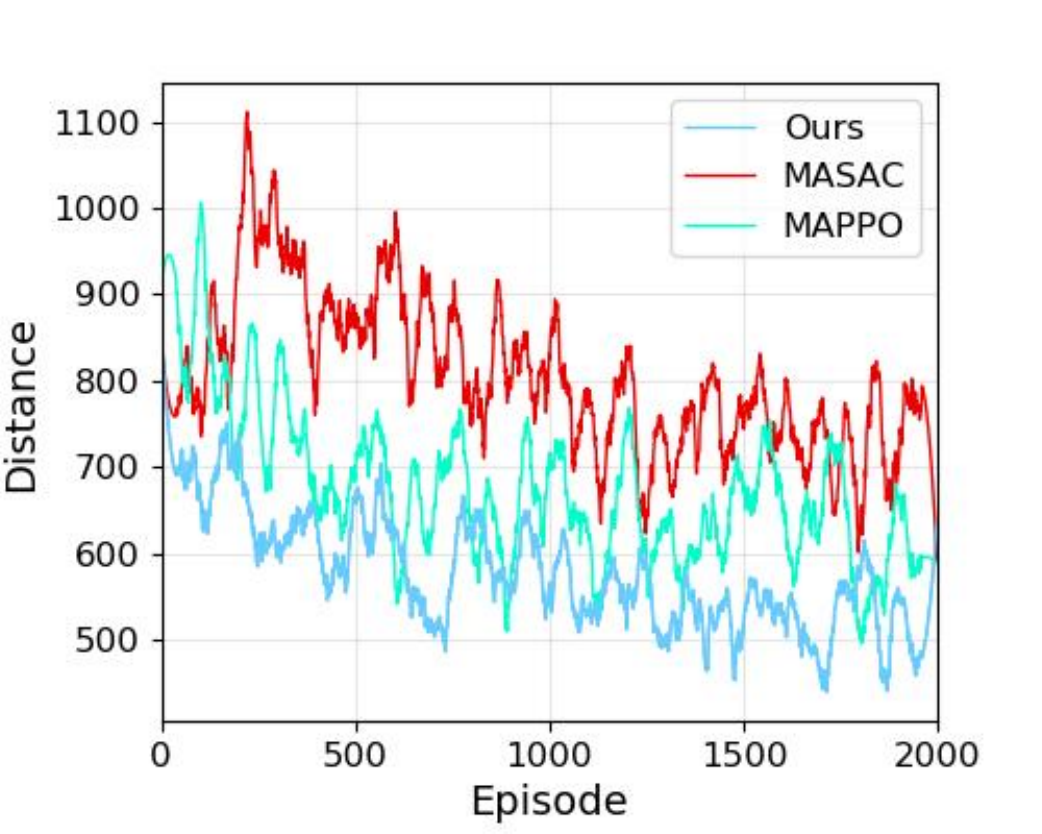}\label{DISTANCEr1}}
\hfil
\subfloat[5 agents with 40 tasks]{\includegraphics[width=0.33\textwidth]{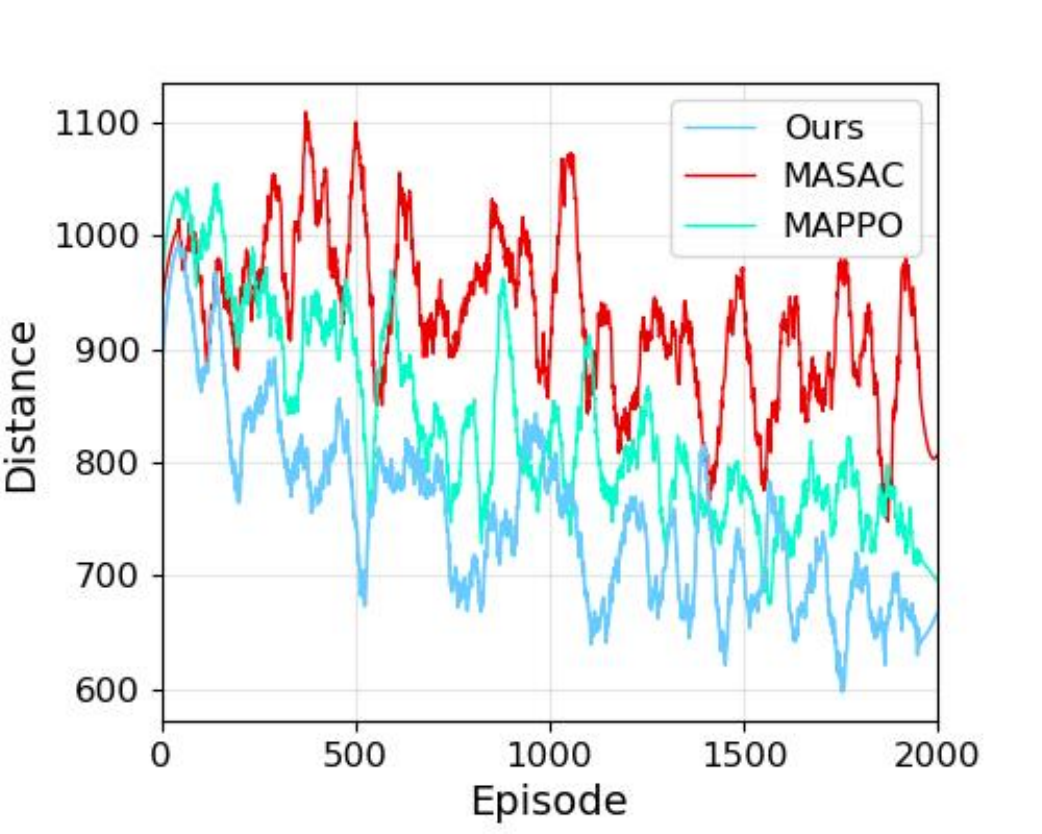}\label{DISTANCEr2}}
\hfil
\subfloat[5 agents with 60 tasks]{\includegraphics[width=0.33\textwidth]{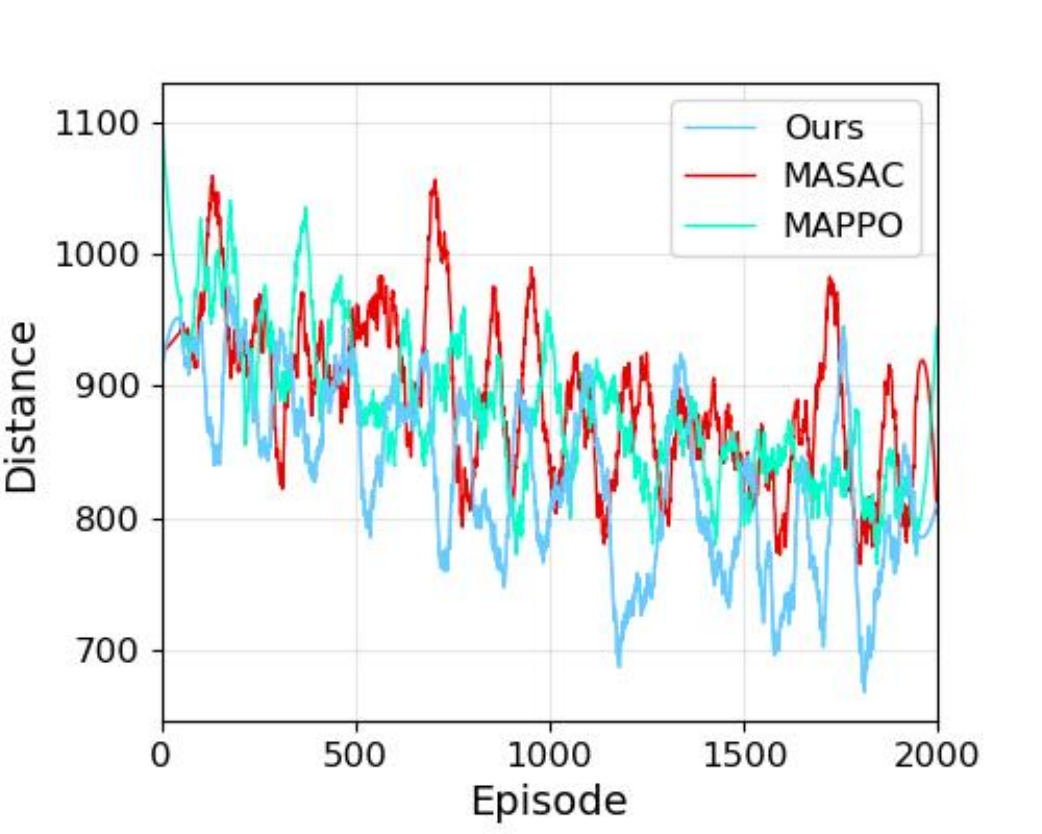}\label{DISTANCEr3}}
\\
\subfloat[10 agents with 20 tasks]{\includegraphics[width=0.33\textwidth]{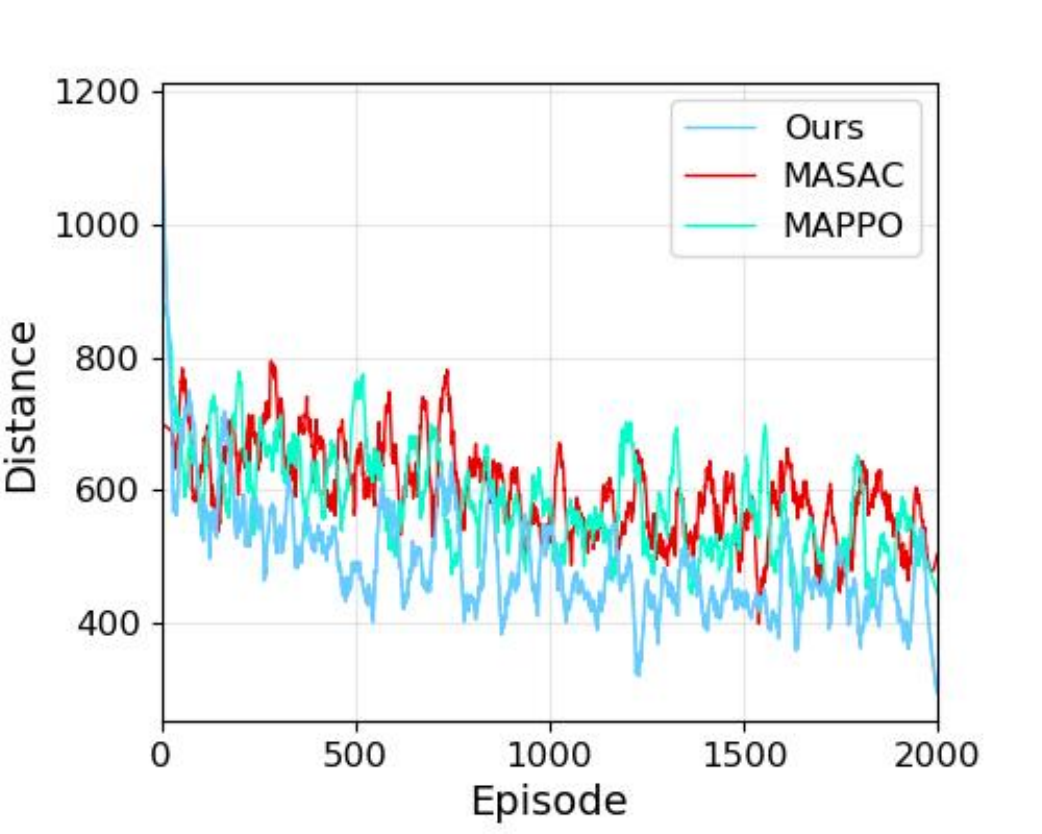}\label{DISTANCEr4}}
\hfil
\subfloat[10 agents with 40 tasks]{\includegraphics[width=0.33\textwidth]{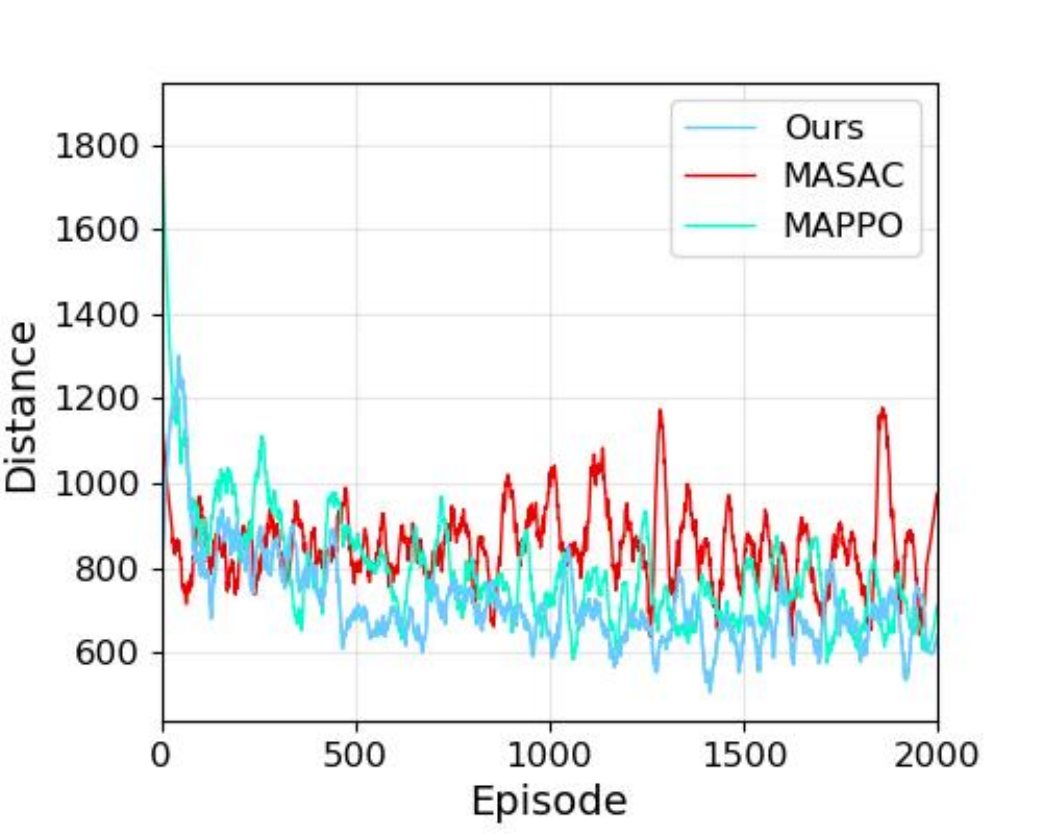}\label{DISTANCEr5}}
\hfil
\subfloat[10 agents with 60 tasks]{\includegraphics[width=0.33\textwidth]{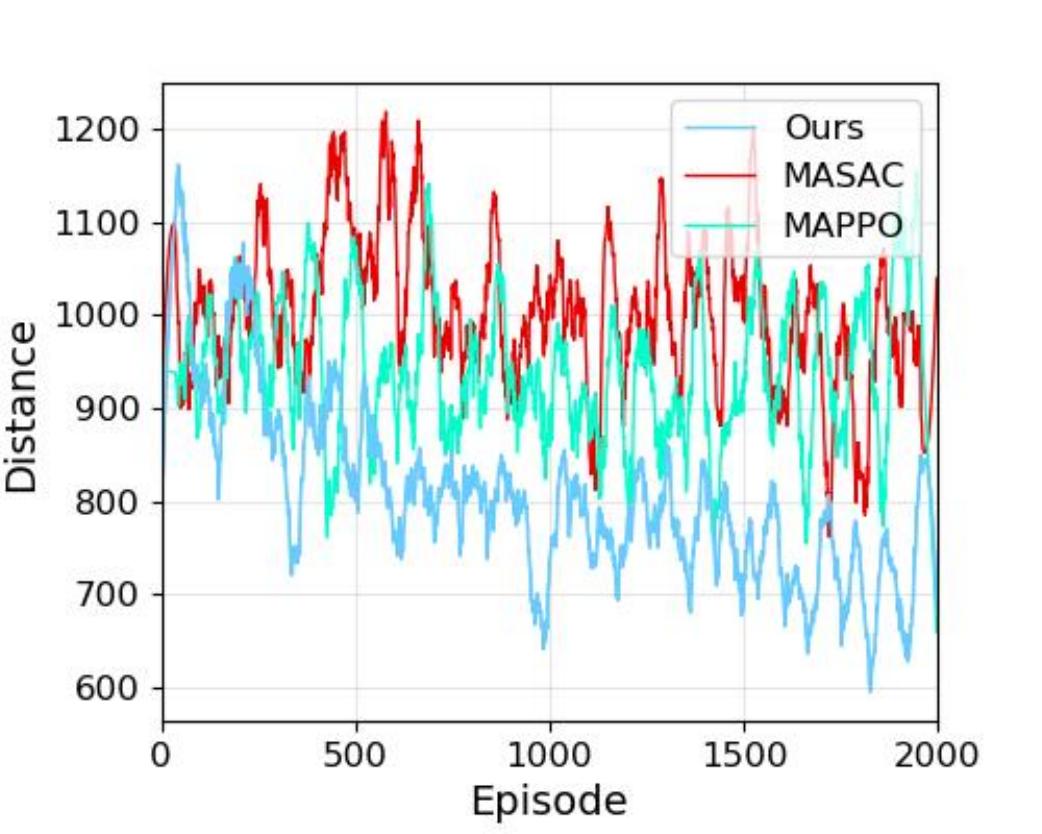}\label{DISTANCEr6}}
\\
\subfloat[20 agents with 20 tasks]{\includegraphics[width=0.33\textwidth]{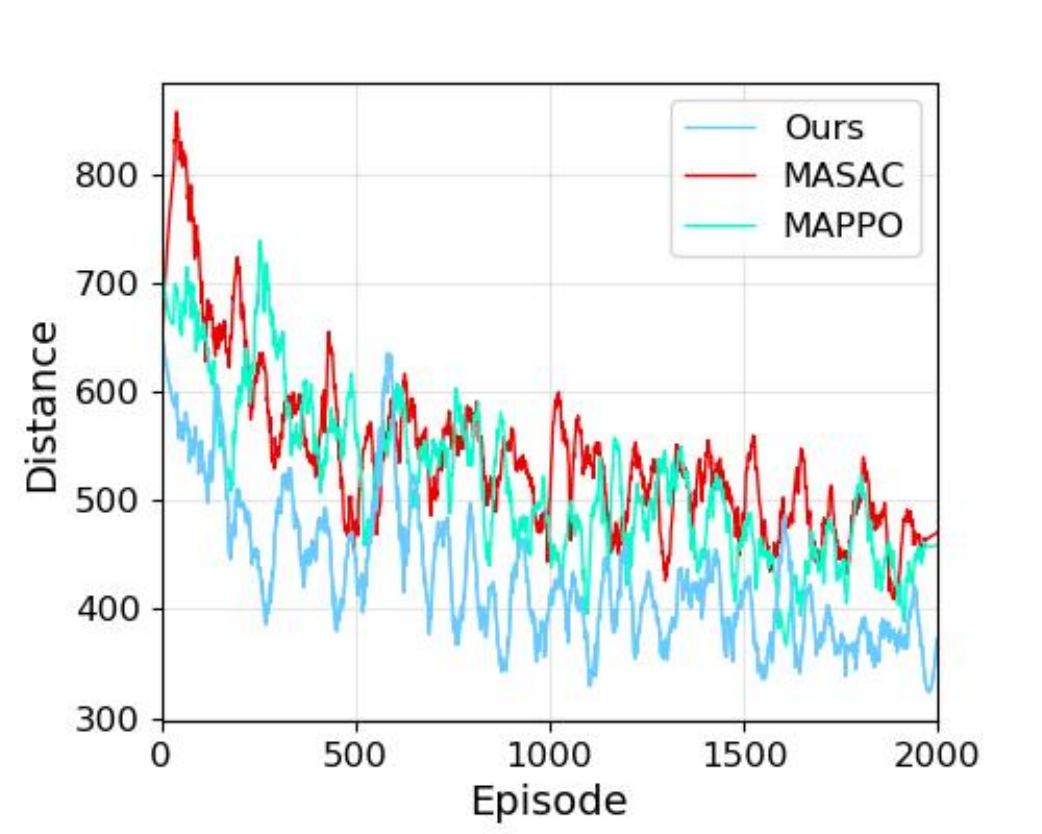}\label{DISTANCEr7}}
\hfil
\subfloat[20 agents with 40 tasks]{\includegraphics[width=0.33\textwidth]{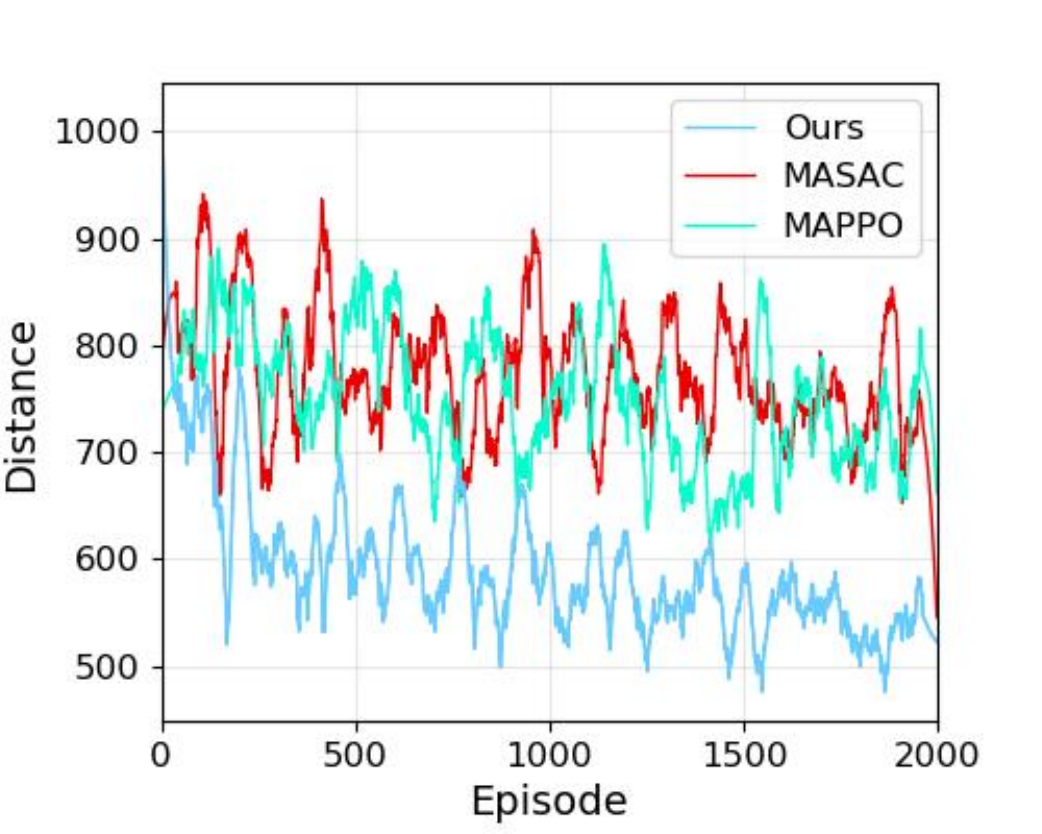}\label{DISTANCEr8}}
\hfil
\subfloat[20 agents with 60 tasks]{\includegraphics[width=0.33\textwidth]{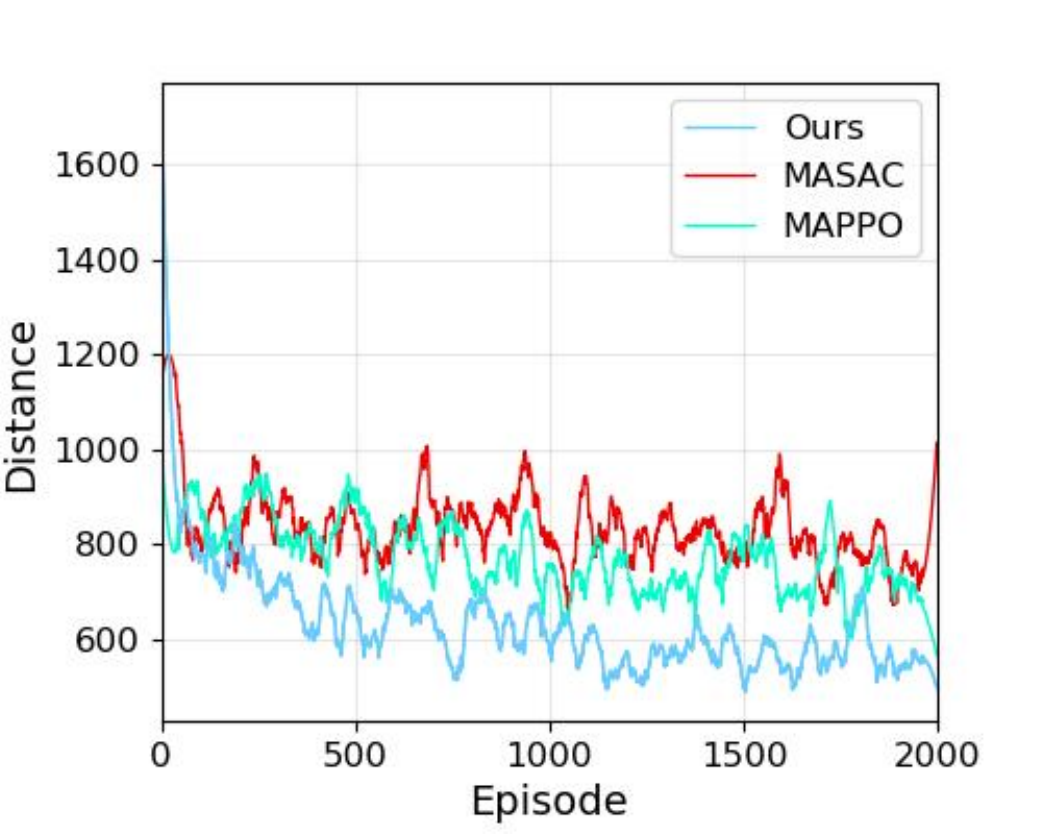}\label{DISTANCEr9}}

\caption{Performance of total travel distance in different scenarios.}
\label{fig_DISTANCE}
\end{figure*}

Fig.~\ref{fig_TIMESTEP} presents the performance of time-step consumption under different agent-task configurations. A consistent downward trend can be observed, and the proposed method generally achieves lower time-step requirements than MASAC and MAPPO across all scenarios. This indicates that tasks are completed more efficiently, with fewer simulation steps needed to reach termination. For instance, in the case of 10 agents and 60 tasks, the proposed method converges rapidly to around 80 steps, while MASAC and MAPPO remain at higher levels with larger fluctuations. The reduced consumption reflects accelerated coordination among agents and suggests that the IRL-based reward shaping encourages policies that prioritize timely completion in addition to energy efficiency.  

Table~\ref{all_table} reports the execution-phase performance of the three algorithms across different agent–task scales. The results demonstrate that the proposed method consistently achieves superior performance in terms of cumulative rewards, timestep consumption, and total travel distance, with most values either being the best or statistically indistinguishable from the best within confidence intervals. This validates the effectiveness of the learned policy beyond the training phase and underlines its robustness when deployed.  

For light-load scenarios such as 5 agents with 20 tasks, the proposed method yields notably higher cumulative rewards (48.31 vs. 42.43 for MAPPO and 30.06 for MASAC) while simultaneously reducing timesteps and travel distance. This indicates that even in relatively uncongested conditions, the attention-augmented IRL framework encourages policies that complete tasks faster and with fewer redundant movements.  

In heavy-load settings such as 10 agents with 60 tasks, the advantage becomes more pronounced. The proposed approach attains 196.91 cumulative rewards compared to 169.53 and 144.57 for MAPPO and MASAC, respectively. It also reduces the timestep consumption to 89.90, while both baselines remain above 100. The lower travel distance further confirms improved spatial efficiency. These outcomes suggest that reward shaping combined with temporal–structural attention allows agents to coordinate more effectively under congestion, mitigating delays and avoiding unnecessary trajectories.

The advantage is also evident in lighter-load scenarios such as 20 agents with 20 tasks. Even though all methods achieve relatively low step counts, the proposed approach consistently reaches a smaller steady-state value and exhibits greater stability. This outcome highlights its robustness across scales: the integration of temporal features through MHSA and structural dependencies through GAT allows agents to anticipate future interactions. This capability helps them avoid redundant movements and thereby reduce overall task duration.

Fig.~\ref{fig_DISTANCE} presents the performance of total travel distance in different agent-task configurations. The proposed method consistently produces shorter trajectories than MASAC and MAPPO, which indicates that agents learn to minimize redundant movements during task execution. In the scenario with 10 agents and 40 tasks, the reduction is particularly pronounced: the distance converges to around 700, while the baselines fluctuate near 800 or higher. The smaller distance implies more efficient spatial coordination, as agents distribute themselves adaptively across tasks rather than overlapping routes. This effect can be ascribed to the attention-guided reward inference, which provides trajectory-level features that penalize unnecessary motion while preserving task efficiency.  

In contrast, under lighter loads such as 20 agents with 20 tasks, the differences between methods are less dramatic but remain consistent. The proposed approach still converges to a lower steady-state distance with reduced oscillation. The robustness across scales suggests that the integration of temporal and structural attention allows agents to anticipate spatial conflicts and avoid excessive path overlap. By incorporating expert-guided adversarial training, the policy generalizes to both congested and sparse conditions, yielding travel patterns that balance task completion and energy conservation.

\begin{table}[width=1.0\linewidth,cols=6,pos=t]
\caption{Performance Evaluation Under Different Scenarios in the Execution Phase}
\label{all_table}

\begin{tabular*}{\tblwidth}{@{} C C L C C C @{}}
\toprule
\textbf{Number of Agents} & 
\textbf{Number of Tasks} & 
\textbf{Algorithm} & 
\textbf{Cumulative Rewards} & 
\textbf{Timestep Consumption} & 
\textbf{Total Travel Distance} \\
\midrule

\multirow{9}{*}{5} 
  & \multirow{3}{*}{20} & Ours  & \textbf{48.31$\pm$3.38} & \textbf{105.27$\pm$10.99} & \textbf{520.32$\pm$40.95} \\
  &                     & MAPPO & 42.43$\pm$2.09 & 146.26$\pm$13.48 & 602.92$\pm$55.70 \\
  &                     & MASAC & 30.06$\pm$1.88 & 161.47$\pm$17.28 & 749.13$\pm$66.70 \\
\cmidrule{2-6}

  & \multirow{3}{*}{40} & Ours  & \textbf{126.89$\pm$3.19} & \textbf{164.20$\pm$13.74} & \textbf{659.45$\pm$45.37} \\
  &                     & MAPPO & 103.56$\pm$2.96 & 173.01$\pm$10.99 & 720.09$\pm$54.94 \\
  &                     & MASAC & 96.88$\pm$2.40 & 188.27$\pm$13.77 & 884.34$\pm$65.72 \\
\cmidrule{2-6}

  & \multirow{3}{*}{60} & Ours  & \textbf{204.99$\pm$3.39} & \textbf{180.88$\pm$11.66} & \textbf{798.18$\pm$65.02} \\
  &                     & MAPPO & 174.71$\pm$3.07 & 203.56$\pm$15.78 & 823.11$\pm$57.23 \\
  &                     & MASAC & 153.32$\pm$2.77 & 218.31$\pm$14.18 & 845.96$\pm$64.03 \\
\midrule

\multirow{9}{*}{10} 
  & \multirow{3}{*}{20} & Ours  & \textbf{56.60$\pm$3.28} & \textbf{80.19$\pm$7.06} & \textbf{426.71$\pm$42.12} \\
  &                     & MAPPO & 49.36$\pm$2.55 & 111.04$\pm$8.59 & 521.16$\pm$47.84 \\
  &                     & MASAC & 42.45$\pm$2.44 & 119.33$\pm$9.56 & 526.57$\pm$49.36 \\
\cmidrule{2-6}

  & \multirow{3}{*}{40} & Ours  & \textbf{130.98$\pm$3.85} & \textbf{74.06$\pm$5.89} & \textbf{646.85$\pm$50.31} \\
  &                     & MAPPO & 104.51$\pm$3.29 & 102.99$\pm$15.07 & 682.33$\pm$62.23 \\
  &                     & MASAC & 95.47$\pm$3.54 & 114.90$\pm$8.69 & 808.57$\pm$80.18 \\
\cmidrule{2-6}

  & \multirow{3}{*}{60} & Ours  & \textbf{196.91$\pm$5.56} & \textbf{89.90$\pm$8.69} & \textbf{746.33$\pm$59.03} \\
  &                     & MAPPO & 169.53$\pm$4.73 & 104.44$\pm$10.01 & 970.80$\pm$106.81 \\
  &                     & MASAC & 144.57$\pm$5.42 & 117.79$\pm$12.60 & 944.61$\pm$92.04 \\
\midrule

\multirow{9}{*}{20} 
  & \multirow{3}{*}{20} & Ours  & \textbf{56.92$\pm$2.53} & \textbf{14.31$\pm$1.61} & \textbf{371.03$\pm$26.01} \\
  &                     & MAPPO & 55.31$\pm$1.85 & 19.93$\pm$2.68 & 441.27$\pm$27.96 \\
  &                     & MASAC & 46.53$\pm$1.43 & 33.84$\pm$3.18 & 470.13$\pm$31.13 \\
\cmidrule{2-6}

  & \multirow{3}{*}{40} & Ours  & \textbf{125.92$\pm$4.41} & \textbf{42.67$\pm$6.11} & \textbf{541.84$\pm$36.64} \\
  &                     & MAPPO & 114.24$\pm$3.15 & 54.44$\pm$4.98 & 724.26$\pm$55.86 \\
  &                     & MASAC & 92.74$\pm$3.23 & 62.90$\pm$5.43 & 696.17$\pm$50.32 \\
\cmidrule{2-6}

  & \multirow{3}{*}{60} & Ours  & \textbf{203.85$\pm$6.44} & \textbf{66.98$\pm$11.45} & \textbf{538.14$\pm$36.59} \\
  &                     & MAPPO & 178.61$\pm$4.02 & 67.91$\pm$8.79 & 665.58$\pm$38.11 \\
  &                     & MASAC & 155.51$\pm$2.17 & 83.33$\pm$12.74 & 785.39$\pm$63.38 \\
\bottomrule
\end{tabular*}
\end{table}

A similar trend is observed for large-scale configurations such as 20 agents with 40 or 60 tasks. Despite the enlarged action space and dense interaction graph, the proposed method maintains the highest rewards and the lowest or near-lowest cost metrics. The fact that MAPPO occasionally approaches the proposed method in single indicators, such as timestep consumption for 20 agents and 60 tasks, highlights that MAPPO benefits from stable updates. However, it still lacks the trajectory-level efficiency introduced by IRL-based inference.

\begin{figure*}
\centering
\subfloat[5 agents with 20 tasks]{\includegraphics[width=0.33\textwidth]{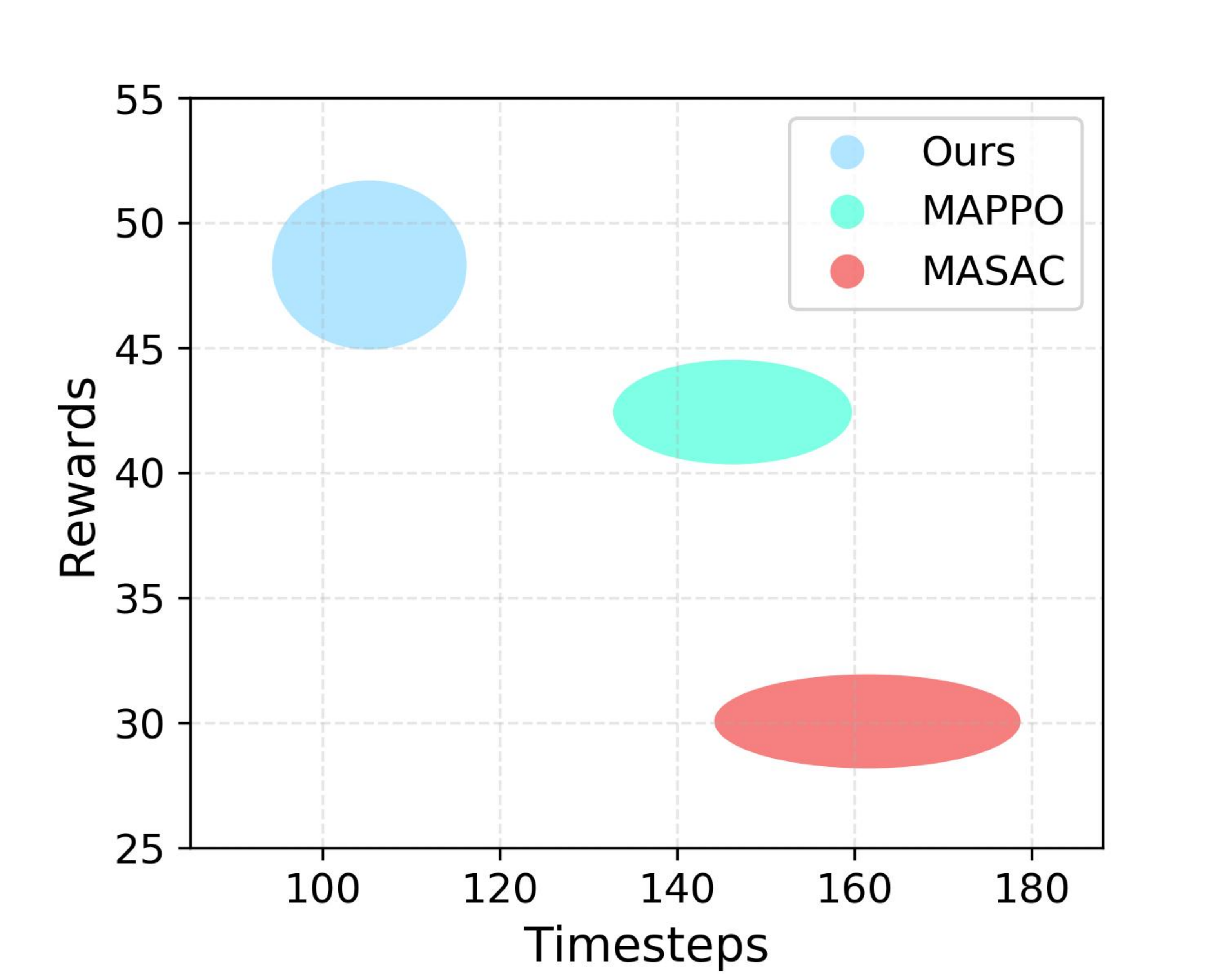}\label{Timestep-Rewardsr1}}
\hfil
\subfloat[5 agents with 40 tasks]{\includegraphics[width=0.33\textwidth]{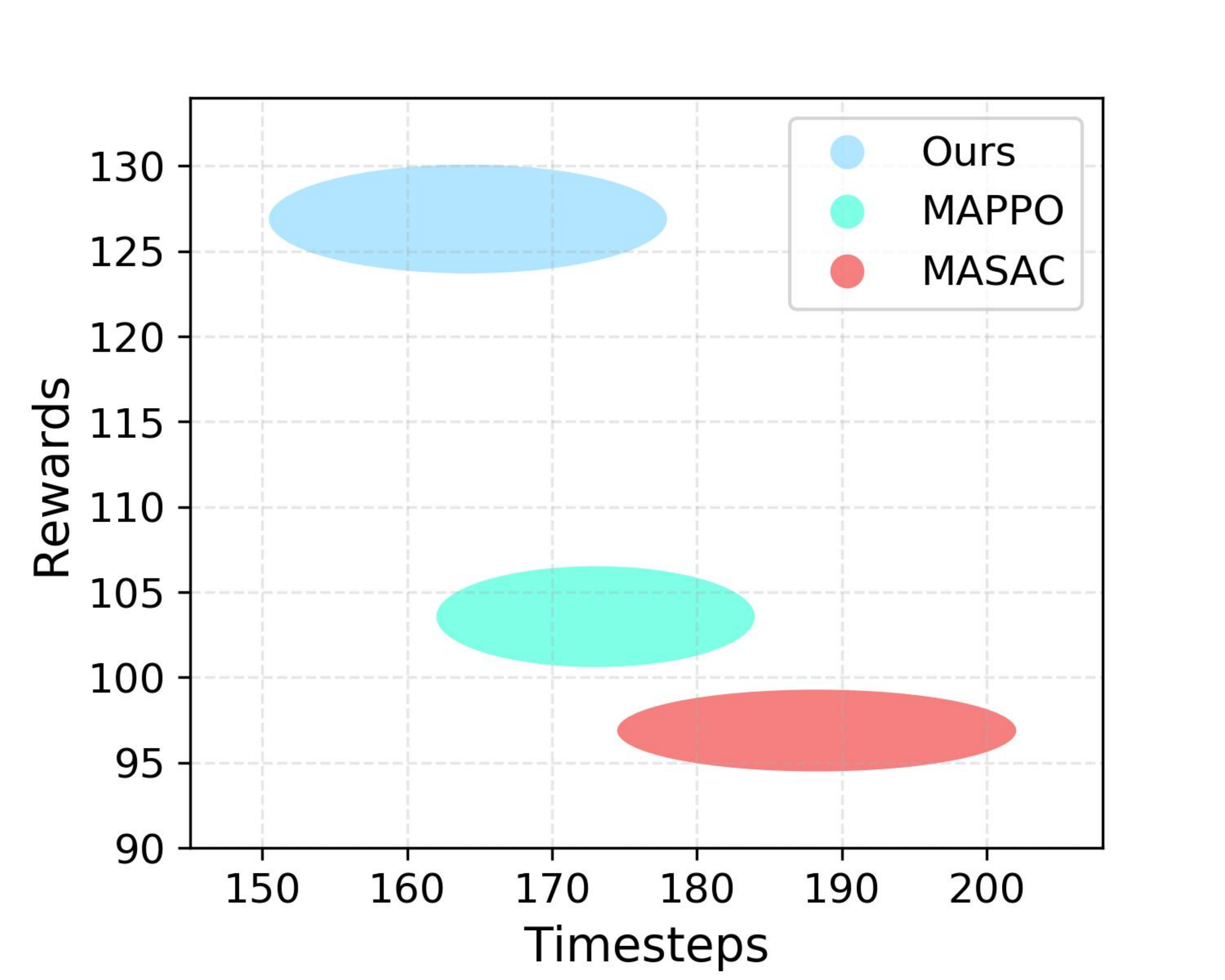}\label{Timestep-Rewardsr2}}
\hfil
\subfloat[5 agents with 60 tasks]{\includegraphics[width=0.33\textwidth]{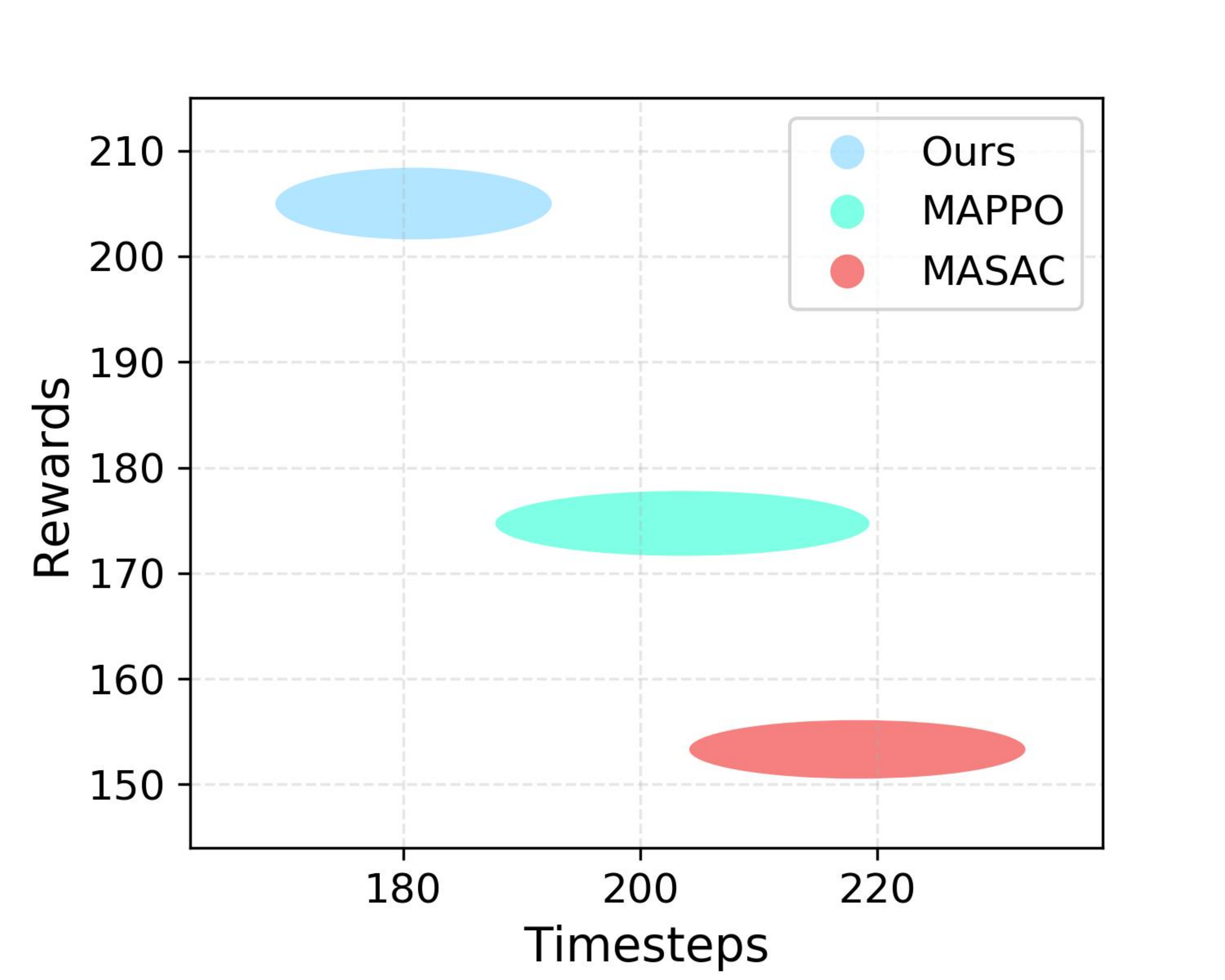}\label{Timestep-Rewardsr3}}
\\
\subfloat[10 agents with 20 tasks]{\includegraphics[width=0.33\textwidth]{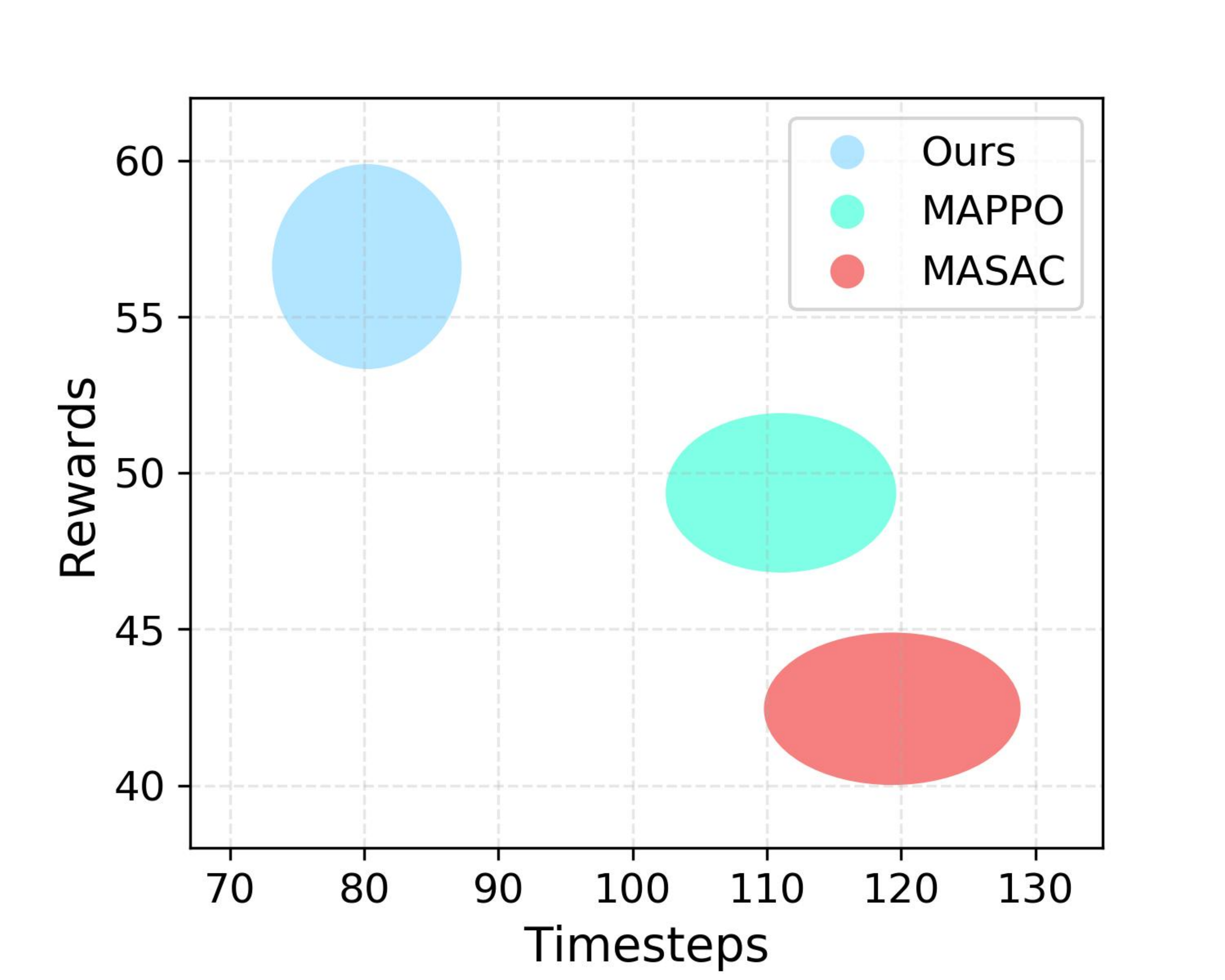}\label{Timestep-Rewardsr4}}
\hfil
\subfloat[10 agents with 40 tasks]{\includegraphics[width=0.33\textwidth]{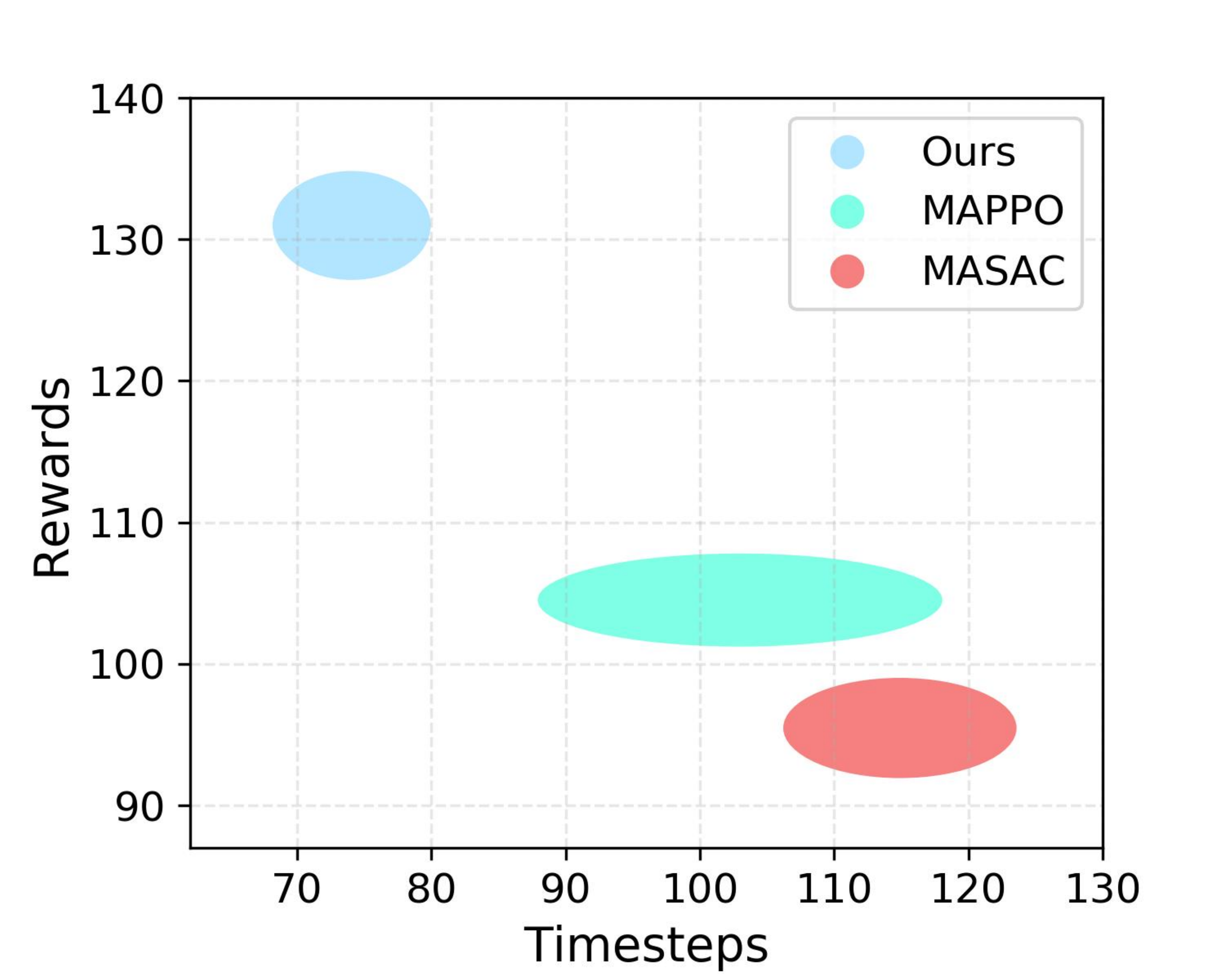}\label{Timestep-Rewardsr5}}
\hfil
\subfloat[10 agents with 60 tasks]{\includegraphics[width=0.33\textwidth]{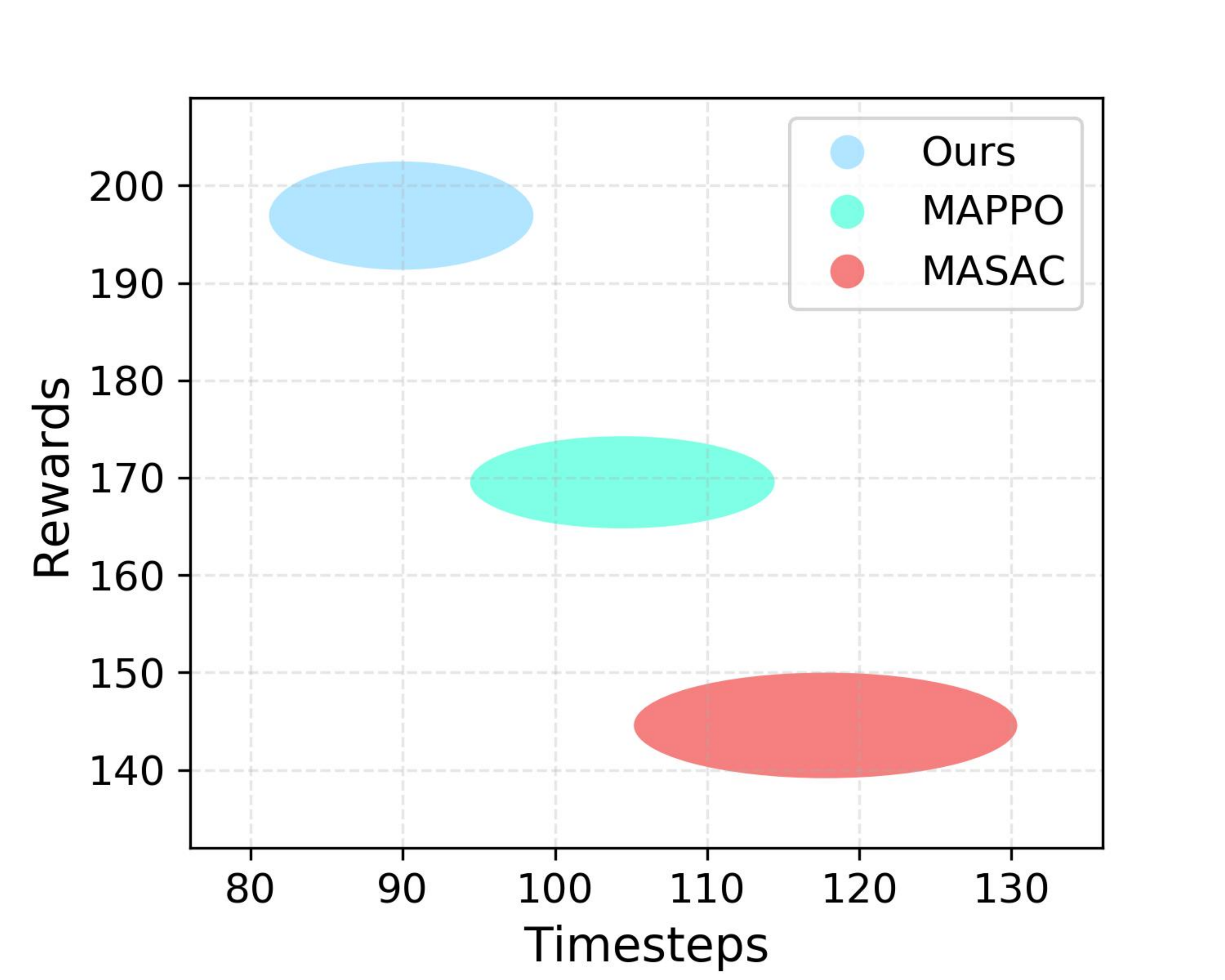}\label{Timestep-Rewardsr6}}
\\
\subfloat[20 agents with 20 tasks]{\includegraphics[width=0.33\textwidth]{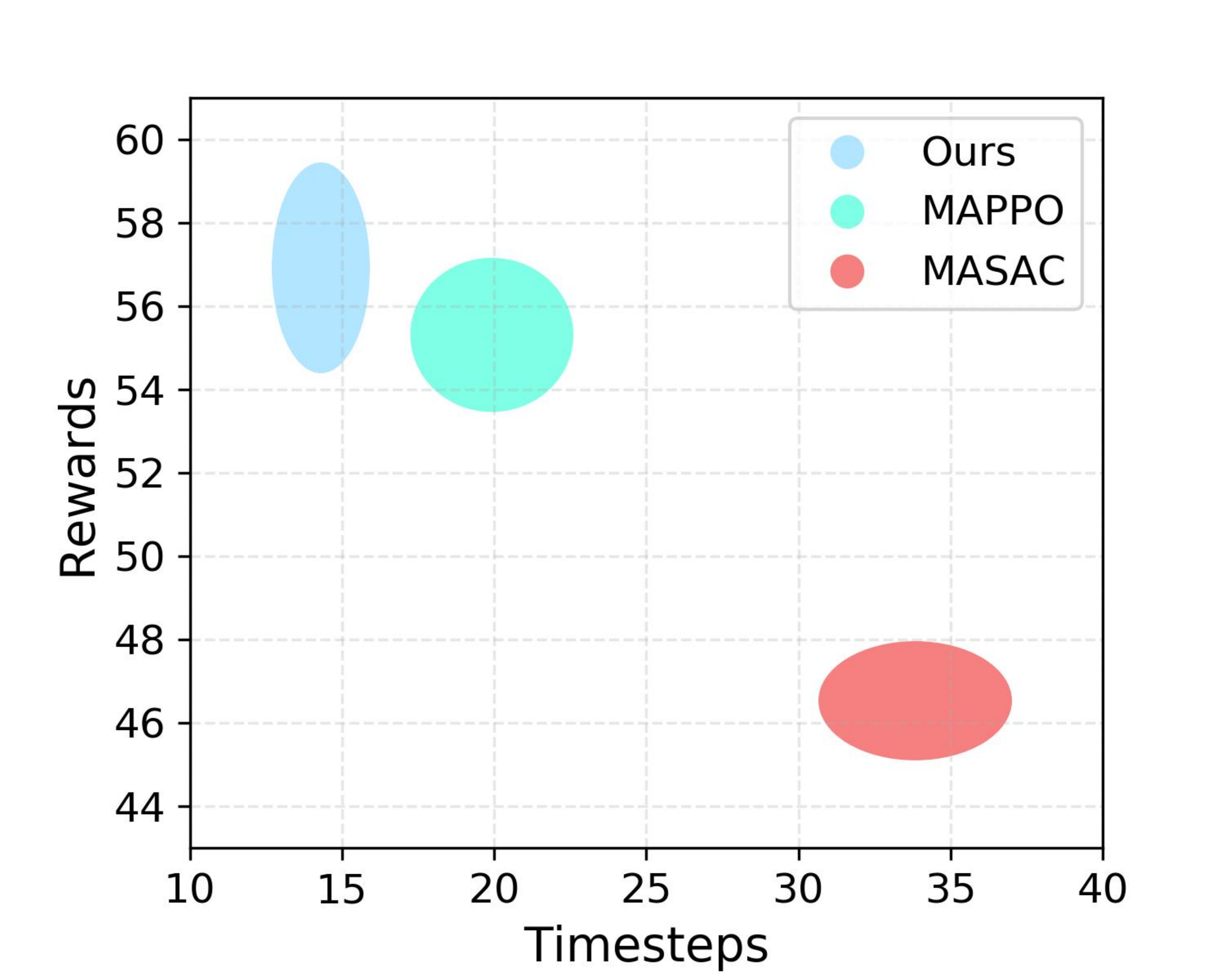}\label{Timestep-Rewardsr7}}
\hfil
\subfloat[20 agents with 40 tasks]{\includegraphics[width=0.33\textwidth]{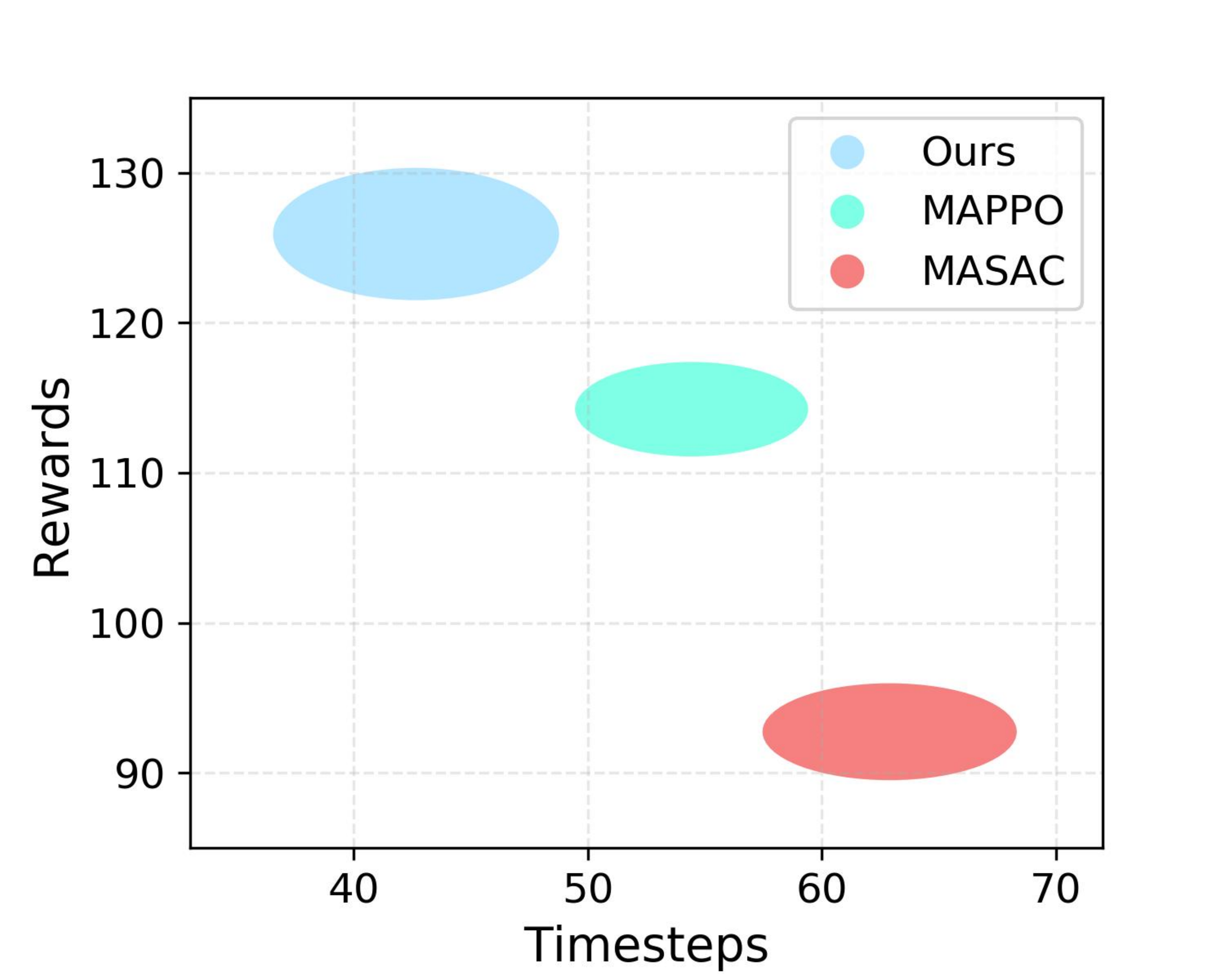}\label{Timestep-Rewardsr8}}
\hfil
\subfloat[20 agents with 60 tasks]{\includegraphics[width=0.33\textwidth]{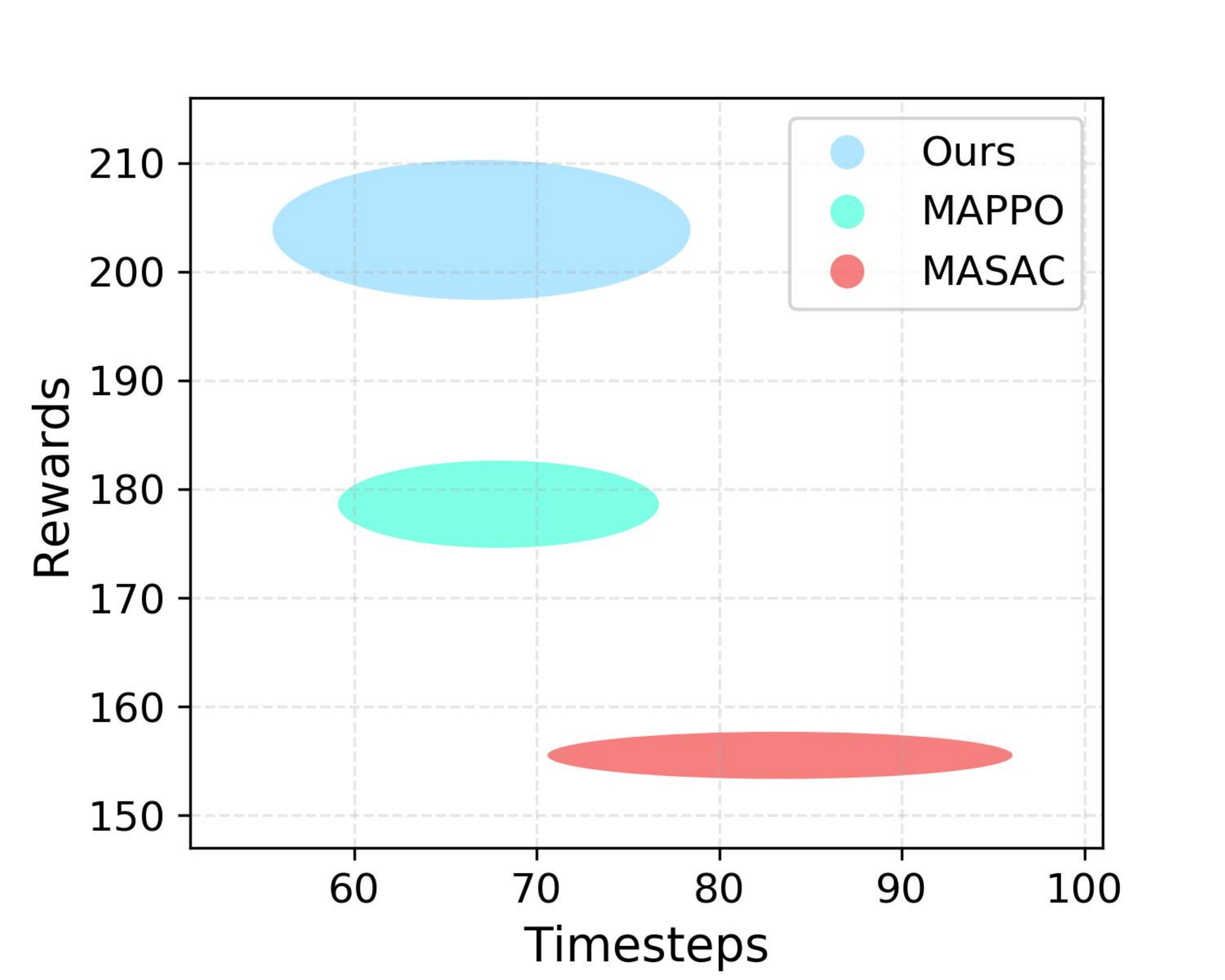}\label{Timestep-Rewardsr9}}

\caption{Relationship between rewards and timesteps in different scenarios.}
\label{fig_Timestep-Rewards}
\end{figure*}

Fig.~\ref{fig_Timestep-Rewards} presents the relationship between cumulative rewards and timestep consumption across all nine agent-task configurations. A consistent pattern is observed throughout the subfigures: the proposed method achieves the highest reward while requiring the fewest timesteps, with MAPPO occupying the middle range and MASAC showing both lower rewards and higher timestep consumption. The ellipses of the three algorithms are clearly separated in all scenarios, indicating a stable performance hierarchy and minimal overlap between their confidence regions.

For each fixed number of agents, increasing the task load from 20 to 40 and 60 shifts all methods toward higher rewards and higher timestep usage, reflecting the increased difficulty of coordinating more tasks. Nevertheless, Ours consistently remains the left-most and upper-most ellipse, demonstrating that it completes tasks more efficiently while achieving higher cumulative returns. This trend holds even in more challenging cases such as 10 agents with 60 tasks (Fig.~\ref{Timestep-Rewardsr6}) and 20 agents with 60 tasks (Fig.~\ref{Timestep-Rewardsr9}), where MAPPO and MASAC both exhibit notable increases in timestep consumption.

Across different agent scales, from 5 to 20 agents, the timesteps of all methods generally decrease, as more agents are available to share the workload. However, the relative positions of the three algorithms remain consistent across all subplots: Ours achieves the best reward–timestep trade-off, MAPPO follows with moderate performance, and MASAC consistently lies in the lower-reward and higher-timestep region.

Overall, both the numerical results and the visual analysis consistently demonstrate that the proposed method outperforms MAPPO and MASAC. Higher cumulative rewards indicate that the learned policy better aligns with the joint objective of minimizing energy and completion time. Lower time-step consumption reflects accelerated coordination and improved temporal efficiency. Shorter travel distances highlight enhanced spatial efficiency and reduced redundant motion. Taken together, these findings suggest that the integration of IRL-based reward inference with attention-augmented trajectory representation yields policies that are simultaneously more optimal, more efficient, and more scalable across diverse task loads. Since the environment, hyperparameters, and baseline configurations are kept identical, the consistent gain can be ascribed to the architectural and algorithmic design rather than to differences in experimental settings.

\subsection{Ablation Analysis}

\begin{figure}
\centering
\includegraphics[width=0.8\textwidth]{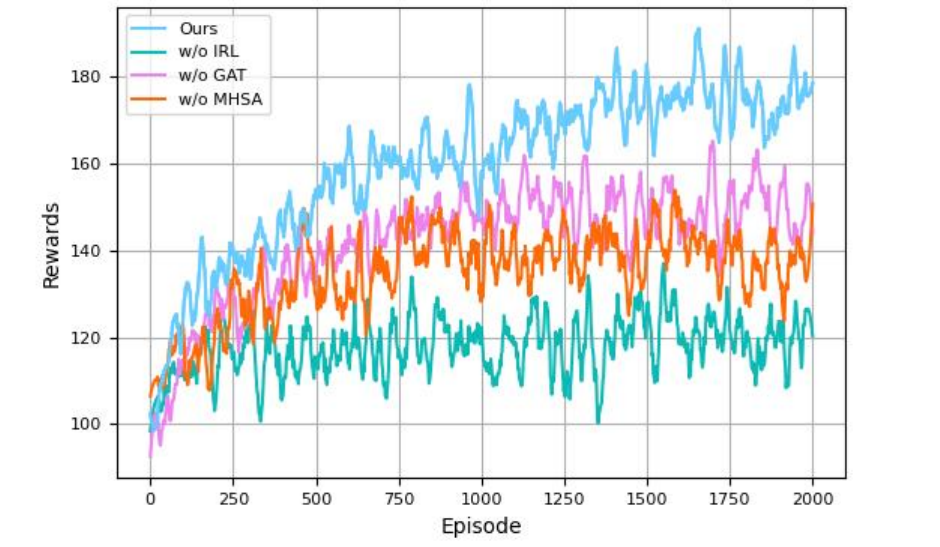}
\caption{Ablation Experiment Results.}\label{abl}
\end{figure}

Fig.~\ref{abl} presents the ablation study, where the proposed model is compared with three simplified variants obtained by removing GAT, MHSA, or the IRL-based reward inference module. The full model consistently achieves higher cumulative rewards with more stable convergence behavior, indicating that each component contributes to the overall performance.

When the graph attention network is removed (w/o GAT), a noticeable degradation is observed in both convergence speed and final performance. This result suggests that explicitly modeling spatial interactions between agents and tasks is important for effective coordination. By enabling adaptive weighting of neighboring tasks and state information, GAT supports informed allocation decisions. In its absence, agents rely primarily on aggregated or locally limited information, which reduces coordination efficiency and leads to suboptimal task assignments.

Removing the multi-head self-attention module (w/o MHSA) results in a further performance decline. MHSA facilitates the modeling of long-range temporal dependencies across inter-task trajectories, providing sequence-level contextual information that supports coherent scheduling over extended horizons. Without this mechanism, policies tend to focus on short-term feedback, leading to less stable training dynamics and degraded long-horizon performance. 

The most notable performance degradation is observed when adversarial reward inference is removed (w/o IRL). In this setting, the reward signal relies solely on handcrafted surrogates, which struggle to balance time efficiency and energy consumption. Consequently, training becomes less stable, performance saturates at substantially lower reward levels, and variance across runs increases. These results indicate that IRL-based reward inference plays a critical role in providing effective reward guidance that aligns policy optimization with implicit expert preferences and the underlying joint objective.  

Overall, the ablation results indicate that each component contributes in a distinct and complementary manner. The GAT module facilitates effective spatial interaction modeling, while MHSA supports the capture of temporal dependencies in inter-task trajectories. The adversarial reward inference mechanism provides adaptive reward guidance that helps balance task efficiency and energy consumption. When combined, these components lead to more stable and scalable policies across diverse agent–task configurations. The results further support the architectural design and highlight the effectiveness of integrating spatial interaction modeling, temporal attention, and adversarial reward inference for addressing the complexity of MATA.

\subsection{Complexity Analysis}
Consider a training scenario with \( n \) agents, \( m \) tasks, and a batch size of \( \mathcal{B} \). In the DRL part, the complexity of the network is linearly related to the number of agents \( n \), batch size \( \mathcal{B} \), network dimensions and the number of layers. Since the network parameters in DRL are fixed, the complexity is \( O(n\mathcal{B}) \). In the generator part, the complexity of MHSA is related to the trajectory length \( L \), feature dimensions, number of attention heads, batch size \( \mathcal{B} \) and the number of layers. With fixed network parameters, the complexity is \( O(\mathcal{B}L^2) \). The complexity of the graph attention network is related to the number of nodes \( n+m \), batch size \( \mathcal{B} \) and feature dimensions. With fixed network parameters, the complexity is \( O(\mathcal{B}(n+m)) \). In the discriminator part, since the network is a simple MLP, the complexity with fixed network parameters is \( O(\mathcal{B}) \). Therefore, the total complexity of the proposed algorithm is \( O(n\mathcal{B} + m\mathcal{B} + \mathcal{B}L^2) \).

\section{Conclusion}
\label{sec:conclusion}
This paper presented an attention-augmented inverse reinforcement learning framework for multi-agent task allocation, where adaptive reward guidance is inferred from expert demonstrations to support coordination under dynamic and non-stationary environments. By integrating temporal trajectory modeling and relational interaction modeling within an adversarial reward inference structure, the proposed approach reduces dependence on manually specified reward functions while providing more stable and informative signals for multi-agent policy optimization. Experimental results across multiple agent–task scales demonstrate consistent improvements over representative MARL baselines in cumulative reward, coordination efficiency, and spatial effectiveness, indicating that structured reward inference can enhance both learning stability and scalability in complex allocation scenarios. Beyond performance gains, the study highlights the role of attention-based spatiotemporal representations as an effective inductive bias for reward learning in multi-agent systems. The proposed framework is modular and extensible, enabling future extensions toward heterogeneous agent capabilities, partially observable environments, and dynamic interaction graphs. Further work will investigate integrating adaptive structure learning and real-world deployment in large-scale intelligent transportation and logistics systems.

\section{Acknowledgment}
This work was supported by the National Natural Science Foundation of China under Grant No.62133011. The authors would like to thank TÜV SÜD for the kind and generous support. We are also grateful for the efforts from our colleagues in Sino German Center of Intelligent Systems.

\section*{Declaration of generative AI and AI-assisted technologies in the manuscript preparation process}
During the preparation of this work, the authors used ChatGPT and Gemini to improve language clarity and grammar. After using these tools, the authors reviewed and edited the content as needed and take full responsibility for the content of the published article.

\section*{Declaration of Competing Interest}

The authors declare that they have no known competing financial interests or personal relationships that could have appeared to influence the work reported in this paper.




\printcredits

\bibliographystyle{model1-num-names}
\bibliography{references}



\end{document}